\documentclass[letterpaper, 10 pt, conference]{ieeeconf}
\usepackage[T1]{fontenc}

\IEEEoverridecommandlockouts                              % This command is only needed if 
\usepackage{amsmath}
\usepackage{amssymb}
\usepackage{graphicx}
\usepackage{subcaption}
\usepackage{gensymb}
\usepackage{algorithm}
\usepackage{algorithmic}
\floatname{algorithm}{Algorithm}
\usepackage[backref, colorlinks,linkcolor=blue]{hyperref}

\usepackage{booktabs}
\usepackage{makecell}
\usepackage[english]{babel}
\usepackage{caption}
\usepackage{multirow}
\usepackage{siunitx}
\title{\LARGE \bf
SENIOR: Efficient Query Selection and Preference-Guided Exploration in Preference-based Reinforcement Learning
}

\author{Hexian Ni$^{1}$, Tao Lu$^{2}$, Haoyuan Hu$^{1}$, Yinghao Cai$^{2}$, Shuo Wang$^{2}$% <-this % stops a space
\thanks{*This work was supported in part by National Natural Science Foundation of China (Grant Number: 62473366, 62273342), the grant of SiYuan collaborative innovation alliance of artificial intelligence science and technology. (Corresponding author: Tao Lu)}% <-this % stops a space
\thanks{$^{1}$Hexian Ni and Haoyuan Hu are with the State Key Laboratory of Multimodal Artificial Intelligence Systems, Institute of Automation, Chinese Academy of Sciences, Beijing 100190, China, and also with the School of Artificial Intelligence, University of Chinese Academy of Sciences, Beijing 100049, China.
        {\tt\small \{nihexian2023, huhaoyuan2025\}@ia.ac.cn}}%
\thanks{$^{2}$Tao Lu, Yinghao Cai and Shuo Wang are with the State Key Laboratory of Multimodal Artificial Intelligence Systems, Institute of Automation, Chinese Academy of
Sciences, Beijing 100190, China.
        {\tt\small \{tao.lu, yinghao.cai, shuo.wang\}@ia.ac.cn}
        }%
}

\begin{document}

\maketitle
\thispagestyle{empty}
\pagestyle{empty}

%%%%%%%%%%%%%%%%%%%%%%%%%%%%%%%%%%%%%%%%%%%%%%%%%%%%%%%%%%%%%%%%%%%%%%%%%%%%%%%%
\begin{figure*}[htbp]
    \centering
    \includegraphics[width=0.9\textwidth]{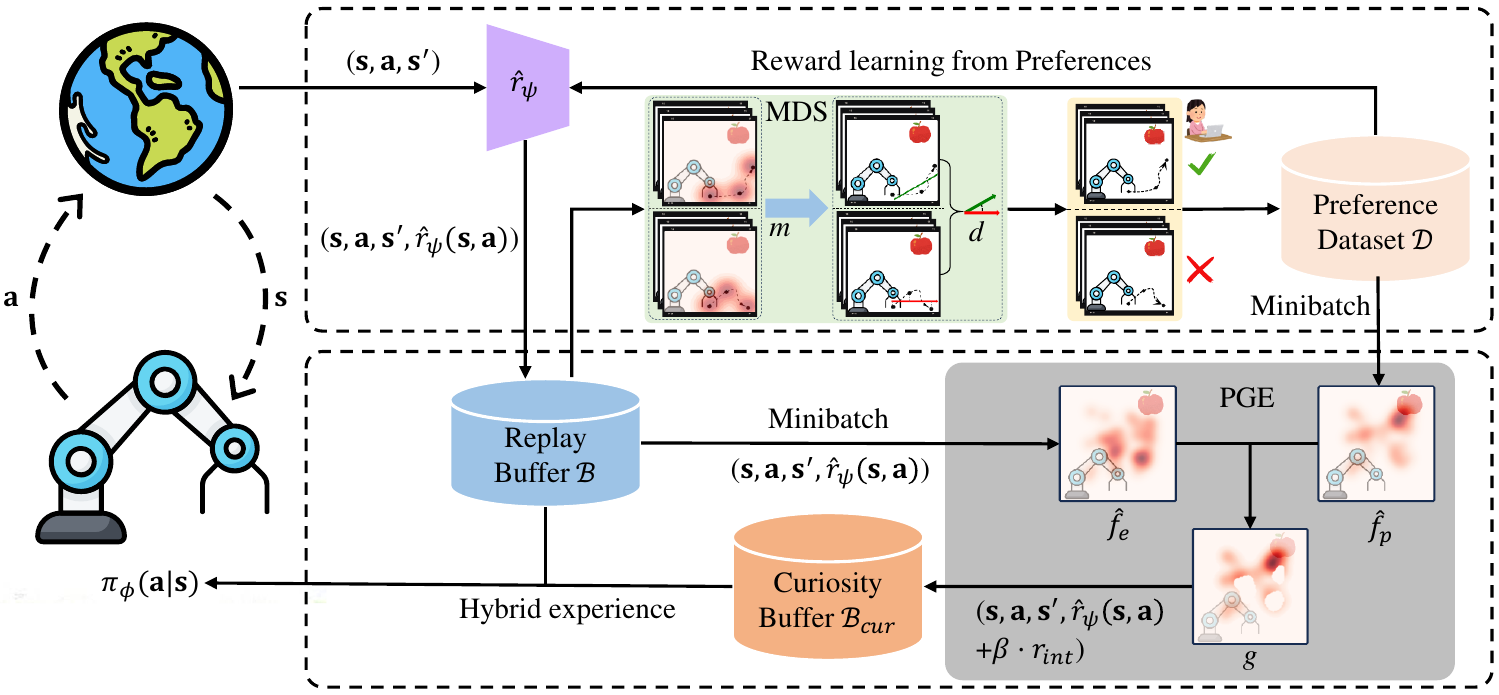}
    \caption{
    Illustration of SENIOR. PGE assigns high task rewards for fewer visits and human-preferred states to encourage efficient exploration through hybrid experience updating policy, which will provide query selection for more valuable task-relevant segments. MDS select easily comparable and meaningful segment pairs with apparent motion distinction for high-quality labels to facilitate reward learning, providing the agent with accurate rewards guidance for PGE exploration. During training, MDS and PGE interact and complement each other, improving both feedback- and exploration-efficiency of PbRL.
    }
    \label{overview of SENIOR.}
\end{figure*}
\begin{abstract}
Preference-based Reinforcement Learning (PbRL) methods provide a solution to avoid reward engineering by learning reward models based on human preferences. However, poor feedback- and sample- efficiency still remain the problems that hinder the application of PbRL. In this paper, we present a novel efficient query selection and preference-guided exploration method, called SENIOR, which could select the meaningful and easy-to-comparison behavior segment pairs to improve human feedback-efficiency and accelerate policy learning with the designed preference-guided intrinsic rewards. Our key idea is twofold: (1) We designed a Motion-Distinction-based Selection scheme (MDS). It selects segment pairs with apparent motion and different directions through kernel density estimation of states, which is more task-related and easy for human preference labeling; (2) We proposed a novel preference-guided exploration method (PGE). It encourages the exploration towards the states with high preference and low visits and continuously guides the agent achieving the valuable samples. The synergy between the two mechanisms could significantly accelerate the progress of reward and policy learning. Our experiments show that SENIOR outperforms other five existing methods in both human feedback-efficiency and policy convergence speed on six complex robot manipulation tasks from simulation and four real-worlds. Videos can be found on our project website: \url{https://2025senior.github.io/}
\end{abstract}

%%%%%%%%%%%%%%%%%%%%%%%%%%%%%%%%%%%%%%%%%%%%%%%%%%%%%%%%%%%%%%%%%%%%%%%%%%%%%%%%
\section{INTRODUCTION}

Reinforcement Learning (RL) has achieved great success in solving decision-making tasks, such as gaming AI \cite{gameing_ai1, gameing_ai2, gaming_ai3}, autonomous driving \cite{auto_driving_1, auto_driving_2, auto_driving_3}, robotic manipulation \cite{robot_1, robot_2, robot_3}, etc. The reward function plays a pivotal role in policy learning of RL. As the complexity of the task increases, it becomes difficult and time-consuming to design suitable reward functions carefully. Preference-based Reinforcement Learning (PbRL) is an excellent solution to the problem of reward engineering. Instead of carefully designing a reward function in advance, it uses the human feedback between two behavior segments to learn reward models that match human preferences, thereby guiding the agent to act on human desires.

However, PbRL methods usually suffer from inefficient feedback, i.e., large numbers of meaningless data pairs are selected and hard to be labeled for recovering reward models. To address this problem, various query selection schemes have been proposed, such as entropy-based query selection \cite{c6, pebble}, disagreement-based query selection \cite{c3, c14, mrn}, policy-aligned query selection \cite{hu2023query}, etc. The goal is to select more informative or accountable queries for efficient reward or policy learning. Another line of work focuses on policy learning, including pre-training policy unsupervised to learn diverse behaviors \cite{pebble}, designing bi-level optimization for both reward and policy \cite{mrn}, exploration based on uncertainty of learned reward models \cite{rune}, etc. This kind of methods mainly improve learning efficiency through supplying more diverse samples or optimizing the Q-function online.
While the above carefully designed query schemes or optimizations have a good try in seeking to maximize reward models consistent with human preferences, they still have trouble selecting meaningful segments for easy preference labeling and as one kind of RL, how to make the exploration to be preference related is less focused, which may hinder the policy learning of PbRL, even with feedback labels.

In this paper, we present a novel efficient query Selection and preferENce-guIded explORation (SENIOR) method to improve both feedback- and exploration-efficiency of PbRL in robot manipulation tasks. Our main contribution is twofold: First, we introduce a Motion-Distinction-based Selection scheme (MDS). By evaluating the state density and motion direction of the robots in the segment, easily comparable and meaningful segment pairs with apparent motion distinction will query human preferences for high-quality labels to facilitate reward learning. Second, we design a Preference-Guided Exploration (PGE) mechanism that utilizes human preferences to encourage agents to visit unfamiliar, human-preferred states for more meaningful exploration in the form of intrinsic rewards. Through the synergy between MDS and PGE, our method could significantly accelerate the progress of reward and policy learning. Experiments show that SENIOR outperforms other methods in feedback-efficiency and policy convergence speed in both simulated and real-world complex robot manipulation tasks.

\section{RELATED WORK}

\subsection{Reinforcement Learning from Human Preferences}

Numerous studies have highlighted the critical role of human preferences in reinforcement learning (RL) \cite{c6, pebble, c14, mrn, rune, c13, c22, c21}. However, addressing complex tasks often requires significant human preferences, which could be expensive to collect. Therefore, selecting informative pairs of behavior segments for querying preferences is essential to enhance feedback-efficiency. Existing query selection schemes include disagreement-based query selection \cite{c3}, entropy-based query selection \cite{pebble}, and policy-aligned query selection \cite{hu2023query}.
Disagreement-based query selection samples segments based on reward uncertainty. Entropy-based query selection samples segments with maximum k-NN distance between states to increase state entropy. This two kinds of approaches improve feedback quality by selecting informative segments, and achieve better performance on complex tasks compared to randomly sampling segments \cite{pebble}. 
To solve the query-policy misalignment problem, \cite{hu2023query} proposed QPA method which designed a policy-aligned query selection scheme by sampling segments from the recent experience and a hybrid experience replay mechanism to ensure the policy updated more frequently on the samples human preferred.

Other works improve the performance by policy learning \cite{pebble, mrn, rune, c13, c22, c33}. 
\cite{pebble} proposed PEBBLE which combines unsupervised pre-training and the technique of relabeling experience to improve feedback-efficiency. 
\cite{mrn} designed one bi-level optimization PbRL framework MRN (Meta-Reward-Net). Through updating Q-function and policy via reward models in inner loop and optimizing the reward function according to Q-function performance on preference data in outer loop, MRN exceeds other methods when few preference labels are available.
\cite{rune} incorporates uncertainty from learned reward models as an exploration bonus to achieve high feedback-efficiency.
Besides, some works also focus on expanding preference labels with unlabeled queries \cite{c14, c21, c34}. 
\cite{c14} leverages unlabeled segments and data augmentation to generate new labels, achieving efficient feedback with fewer preferences. 
\cite{c21} incorporates a triplet loss that directly updates the reward model using unlabeled trajectories to improve reward recovery. 
\cite{c34} proposes self-training augmentation to generate pseudo-labels, combined with peer regularization, to solve the similarity trap and obtain confident labels without noise.
In this paper, we evaluate the motion information and direction similarity in trajectory segments and tend to select segment pairs with apparent motion distinction, which are more task-related meaningful and easy for humans to compare. Our method facilitates high-quality labels and feedback-efficiency.

\subsection{Exploration in RL}

The trade-off of exploitation and exploration is pivotal for RL. Much work has shown that appropriate exploration can accelerate policy learning \cite{c9, explore1_greedy, explore3_action, explore4_parameter}. Exploration methods can be categorized into random exploration and intrinsic motivation. Random exploration includes $\epsilon$-greedy \cite{explore1_greedy}, action space noise \cite{explore3_action, explore2_action} and parameter space noise \cite{explore4_parameter, explore3_parameter}. The main ideas of them are to randomly change action output to access unknown states. Intrinsic motivation methods can be broadly categorized as: count-based \cite{c31, c40, c41}, curiosity-based \cite{c15, c16, c2}, and state entropy-based \cite{pebble, c8, c19}. 
Count-based methods usually use a density model to fit pseudo-counts, and intrinsic rewards are high for state spaces with fewer visits. 
Curiosity-based methods calculate the intrinsic rewards by evaluating whether the states are familiar. \cite{c15} generates a curiosity reward based on the agent’s ability to predict the consequences of actions, which encourages the agent to learn skills that might be useful later.
\cite{c16} trains predictive models and motivates the agent to explore by maximizing the disagreement of these models, allowing the agent to learn skills in self-supervised learning without extrinsic rewards.
State entropy-based methods generally maximize state entropy to encourage agents to explore. \cite{c8} aims to learn a single exploratory policy with high state entropy that matches the state marginal distribution to a given target state distribution for faster exploration. \cite{c19} employs random encoders and the k-NN state entropy estimator to facilitate efficient exploration in high-dimensional observation spaces. 
The idea of our method introduces human preferences into intrinsic reward exploration, encouraging agents to visit unfamiliar, human-preferred states. Our method could continuously supply high-valuable states for policy training. 

\section{BACKGROUND}

This section introduces related concepts of Preference-based Reinforcement Learning.

We consider a standard Markov Decision Process (MDP) \cite{c20}. In discrete time, an agent interacts with the environment can be described as (1) at each time step $t$, the agent performs action $\mathbf{a}_t$ based on the current environment state $\mathbf{s}_t$ and policy $\pi_{\phi}(\mathbf{a}_t|\mathbf{s}_t)$, and (2) the environment state shifts to $\mathbf{s}_{t+1}$, and the reward $r(\mathbf{s}_t, \mathbf{a}_t)$ is returned to the agent. The goal of RL is to learn a policy to maximize the cumulative reward of the current state, e.g., $\mathcal{R}_t = \sum_{i=0}^T{\gamma^i}r(\mathbf{s}_{t+i}, \mathbf{a}_{t+i})$, with $t$ denoting the current time step, $T$ denoting the time domain, and $\gamma\in\left[0, 1\right)$ denoting the discount factor.

PbRL utilizes human preference labels between behavior segments of an agent to learn a reward function $\hat{r_\psi}$, which is used for RL policy $\pi_{\phi}$ learning \cite{c3, c28, c23}. Specifically, $\sigma$ denotes a state-action sequence $\left\{{(\mathbf{s}_k, \mathbf{a}_k), \ldots, (\mathbf{s}_{k+H}, \mathbf{a}_{k+H})}\right\}$, which is typically a short part of whole trajectory. The human expert provides a preference $y$ on two segments $(\sigma^0, \sigma^1)$, e.g., $y\in\left\{(0, 1), (1, 0), (0.5, 0.5)\right\}$, indicating $\sigma^1$ preferred, $\sigma^0$ preferred, and incomparable, respectively. Each feedback label is stored as a triple $(\sigma^0, \sigma^1, y)$ in the preference dataset $\mathcal{D}$. Based on the Bradley-Terry model \cite{c1}, the preference prediction are calculated by the learned reward model $\hat{r_\psi}$:
\begin{equation}
    P_{\psi}[\sigma^{1}\succ\sigma^{0}]=\frac{\exp\sum_{t}\hat{r}_{\psi}(\mathbf{s}_{t}^{1},\mathbf{a}_{t}^{1})}{\sum_{i\in\{0,1\}}\exp\sum_{t}\hat{r}_{\psi}(\mathbf{s}_{t}^{i},\mathbf{a}_{t}^{i})}, 
\end{equation}
where $\sigma^1\succ\sigma^0$ refers to the fact that $\sigma^1$ is more consistent with the human expectations than $\sigma^0$. This model assumes that the cumulative reward sum of the segment exponentially determines the probability that a human expert prefers the segment. Learning the reward model becomes a binary classification problem in the case of supervised learning, i.e., keeping the predicted preferences of the model consistent with humans. Thus the reward function $\hat{r}_{\psi}$ parameterized by $\psi$ is updated to minimize the following cross-entropy loss:

\begin{equation}
\begin{aligned}
\mathcal{L}^{\text{Reward}}=&-\mathop{\mathbb{E}}_{(\sigma^{0},\sigma^{1},y){\sim}\mathcal{D}} \Big[y(0)*\log P_{\psi}[\sigma^0\succ\sigma^1]\\
&\qquad\qquad\qquad\quad+y(1)*\log P_{\psi}[\sigma^1\succ\sigma^0]\Big].
\end{aligned}
\label{eq:cross-entropy}
\end{equation}

\section{METHOD}

In this section, we systematically introduce our method SENIOR, including Motion-Distinction-based Selection (MDS) and Preference-Guided Exploration (PGE), which incorporates two ways to improve both the feedback- and exploration-efficiency of PbRL.

\begin{figure}[htbp]
    \centering
    \includegraphics[width=0.5\textwidth]{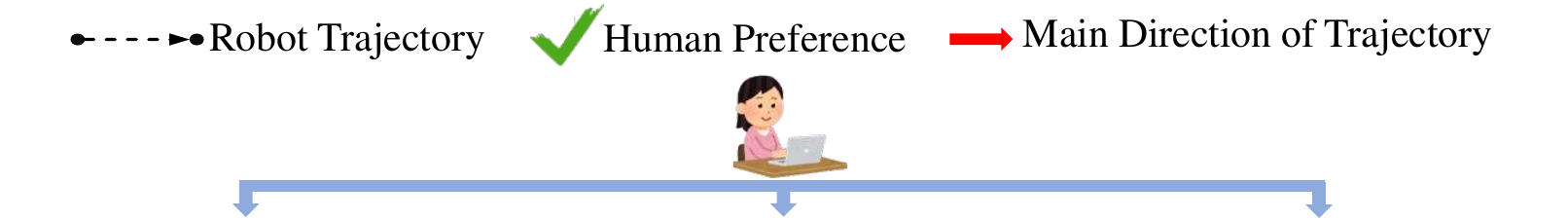}
    \begin{subfigure}[b]{0.155\textwidth}
        \includegraphics[width=\textwidth]{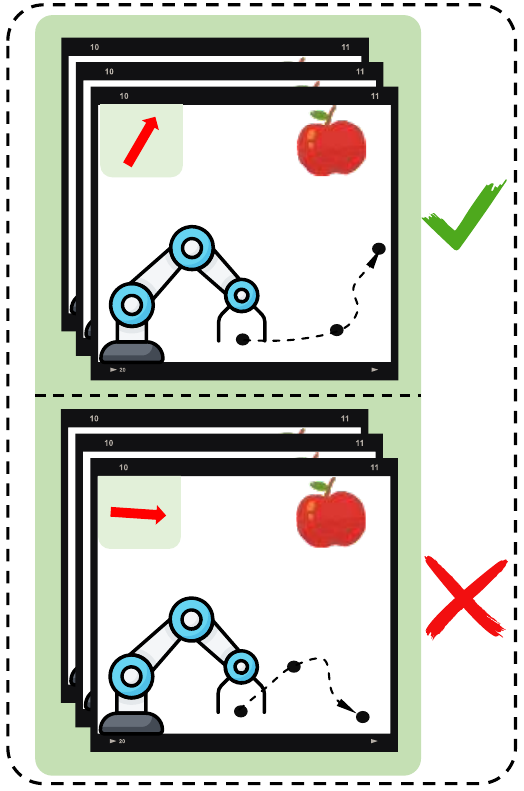}
        \caption{}
        \label{fig1a}
    \end{subfigure}
    \begin{subfigure}[b]{0.155\textwidth}
        \includegraphics[width=\textwidth]{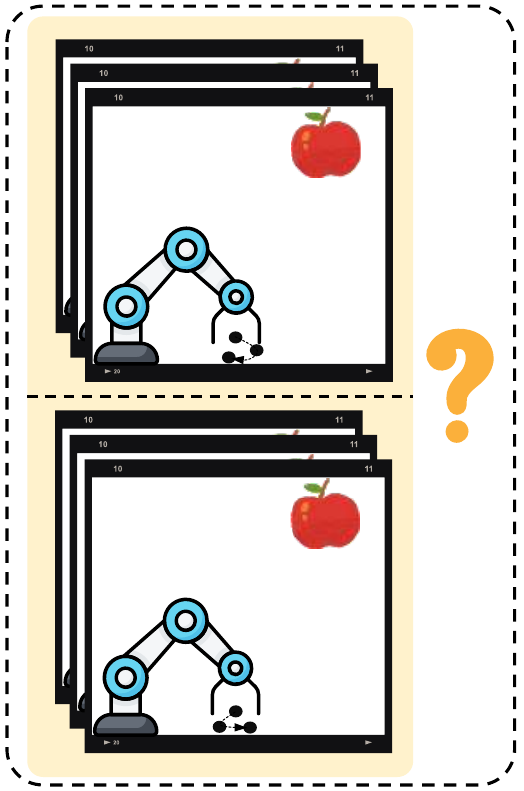}
        \caption{}
        \label{fig1b}
    \end{subfigure}
    \begin{subfigure}[b]{0.155\textwidth}
        \includegraphics[width=\textwidth]{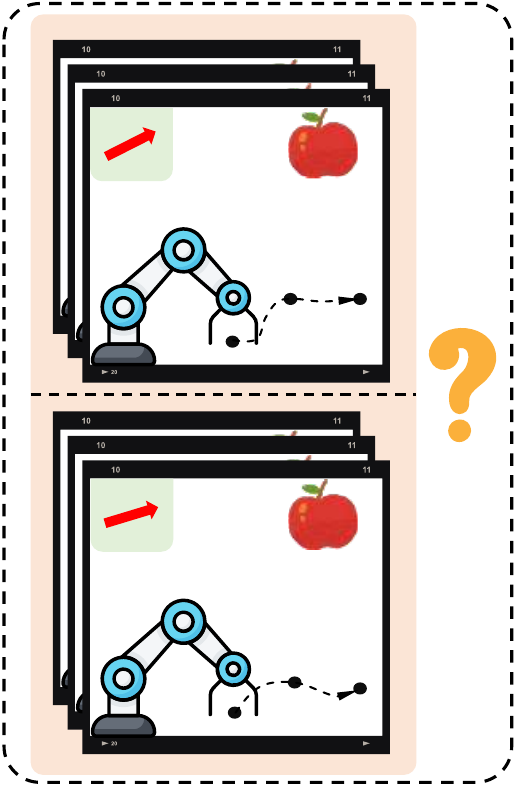}
        \caption{}
        \label{fig1c}
    \end{subfigure}
        \caption{Robot apple grab task. MDS tends to select trajectories that have more motion information and are easy to compare (a) rather than high-density trajectories with less motion information (b) or trajectories with high similarity that are difficult to compare (c).}
    \label{density behavior segments}
    \end{figure}
\subsection{ Motion-Distinction-based Selection (MDS)}

While uncertainty-based (using disagreement or entropy) query selection schemes improve feedback-efficiency \cite{pebble}, they still suffer from the difficulty of selecting meaningful segment pairs for which humans can confidently compare and provide preferences.
For instance, the robot in one segment of the trajectory has no apparent movement. This kind of segment may be selected because of high uncertainty. However, it has less help for task learning and may cause incorrect labels.
So, it is necessary to analyze more motion information in the segment except uncertainty. Based on this idea, we introduce a new query selection scheme MDS. 

In MDS, we first calculate the density of states for each behavior trajectory using Kernel Density Estimation (KDE) \cite{c17}. As shown in Fig.~\ref{density behavior segments}, a high-density trajectory (Fig.~\ref{fig1b}) means the robot always stays around one state, and a low-density trajectory (Fig.~\ref{fig1a}, \ref{fig1c}) means containing more motion information. We tend to select low-density pairs of trajectories. With this principle, we design a motion-score metric $m$ for each segment pair:
\begin{equation}
     m =\frac{1}{\sum_{\mathbf{p}_t\in\sigma^0}\hat{f}_{\sigma^0}(\mathbf{p}_t)+\sum_{\mathbf{p}_t\in\sigma^1}\hat{f}_{\sigma^1}(\mathbf{p}_t)},
     \label{eq:m}
\end{equation}
\begin{equation}
    \hat{f}_{\mathcal{S}}(\mathbf{p_t})=\frac{1}{nh}\sum_{i=1}^{n}K(\frac{\mathbf{p_t}-\mathbf{p}_{i}}{h}), \mathbf{p_i}\in \mathcal{S}, 
    \label{eq:f}
\end{equation}
where $(\sigma^0, \sigma^1)$ denotes a segment pair sampled from replay buffer $\mathcal{B}$, and $\mathbf{p}_t$ denotes the position of the end-effector in $\mathbf{s}_t$. $\hat{f}_{\mathcal{S}}(\cdot)$ denotes an estimation of state density in one segment. $K$ is the Gaussian kernel function, $n$ is the length of the segment, and $h$ is the bandwidth. This process would select segment pairs with high $m$ denoted as $(\sigma^0_m, \sigma^1_m)$.

To further facilitate comparison and achieve more valuable feedback, we emphasize the difference between $\sigma^0_m$ and $\sigma^1_m$. Compared trajectories in Fig.~\ref{fig1a} with that in Fig.~\ref{fig1c}, it becomes easy to label them when their motion directions are distinct, which also brings the result of finding out the task-related segments implicitly. Inspired by this, we design another distinction-score metric $d$ to evaluate the similarity of motion direction $(\mathbf{v}^0_m, \mathbf{v}^1_m)$ between $\sigma^0_m$ and $\sigma^1_m$. 
Here $(\mathbf{v}^0_m, \mathbf{v}^1_m)$ is the eigenvector corresponding to the largest eigenvalue by applying Principal Component Analysis (PCA) to the states within the segment $\sigma^0_m$ and $\sigma^1_m$. The metric $d$ is calculated as cosine similarity:
\begin{equation}
     d =\mathbf{v}^0_m \cdot \mathbf{v}^1_m.
\end{equation}

In general, MDS first randomly samples segment pairs $p$ from the replay buffer $\mathcal{B}$. Then, retains segment pairs $q$ from $p$ with the topest high $m$ score. Finally, obtains per-session segment pairs from $q$ with low $d$. Segment pairs selected by MDS contain more motion information and are easier to compare for humans, thus accelerating reward learning.

\subsection{Preference-Guided Exploration (PGE)}

We propose a novel preference-guided exploration method to provide more diverse and useful segments for the query selection in PbRL. The motivation of PGE is to encourage the exploration of states that humans favor but are unfamiliar to agents. 
In detail, we set up an additional curiosity buffer $\mathcal{B}_{cur}$ and periodically sample data $\mathcal{E}$ from the replay buffer $\mathcal{B}$ to compute exploration KDE $\hat{f}_{\mathcal{E}}$ with Eq. \eqref{eq:f} and add $\mathcal{E}$ to $\mathcal{B}_{cur}$. Through comparing $\hat{f}_{\mathcal{E}}$ with the preference KDE $\hat{f}_{\mathcal{P}}$ calculated
 from the sample data of preference dataset $\mathcal{D}$, we would give higher rewards to the states with high preference density but less visited. The intrinsic reward is calculated by:
\begin{equation}
    r_{int}(\mathbf{p}_i)=\frac{g(\mathbf{p}_i)-\mathop{\min g(\mathbf{p}_i)} \limits_{\mathbf{p}_i\in\mathcal{B}_{cur}}}{\mathop{\max g(\mathbf{p}_i)} \limits_{\mathbf{p}_i\in\mathcal{B}_{cur}}-\mathop{\min g(\mathbf{p}_i)} \limits_{\mathbf{p}_i\in\mathcal{B}_{cur}}},
    \label{eq:r_int}
\end{equation}
\begin{equation}
    g(\mathbf{p}_i)=\frac{\hat{f}_{\mathcal{P}}(\mathbf{p}_i)}{\hat{f}_{\mathcal{E}}(\mathbf{p}_{i})}, 
\end{equation}
Combined with the learned extrinsic rewards $\hat{r}_{\psi}$, we can define the task rewards as:
\begin{equation}
    r_{cur}(\mathbf{s}_i, \mathbf{a}_i)=\hat{r}_{\psi}(\mathbf{s}_i, \mathbf{a}_i)+\beta_t\cdot r_{int}(\mathbf{p}_i), 
    \label{eq:reward_cur}
\end{equation}
where $\beta_t = \beta_{0}(1-\rho)^t$, a hyper-parameter that decays exponentially over time and controls the trade-off between exploration and exploitation of the agent. $\rho$ is the decay rate. 

\section{EXPERIMENTS}
In this section, we first introduce the RL simulation environment for our experiments. Then, our method is compared to five baseline methods in terms of sample- and feedback-efficiency. Ablation experiments show the importance of each component on learning performance. Finally, we show the experimental results on four real-world tasks with a physical robot.

\subsection{Simulation Environment}

We evaluate SENIOR on six robotic continuous control tasks in Meta-World \cite{c24}, including Door Lock, Window Close, Handle Press, Window Open, Door Open and Door Unlock. Similar to previous work \cite{c6, mrn, rune}, to evaluate the tasks quickly, we consider using a scripted teacher that evaluates preferences via the Meta-World environment reward function rather than employing a real human expert. This setup implies that we maintain absolute preferences for those segments that receive higher rewards in the environment. A preference learning approach based on this setup specifies an upper-performance limit, i.e., the agent can directly utilize the environmental reward function for policy learning. In this paper, we selected SAC \cite{sac} method.

SENIOR can be combined with any off-policy PbRL algorithm. To verify the performance, we implement SENIOR on PEBBLE \cite{pebble} (P-SENIOR) and MRN \cite{mrn} (M-SENIOR) and compared our method with five existing typical methods, including PEBBLE, MRN, RUNE \cite{rune}, M-RUNE (RUNE with MRN), QPA \cite{hu2023query}.

\subsection{Implementation Details}
For all methods, we use unsupervised pre-training proposed by PEBBLE and the same hyper-parameters and network architectures as in original papers \cite{mrn, hu2023query, rune}. Similar to RUNE, we fine-tune $\beta_0$ and $\rho$ by setting $\beta_0=0.1$, $\rho\in\left\{0.01, 0.001, 0.0001\right\}$, and report the optimal results. The feedback settings, optimization update frequency and other parameters can be found in the Appendix. 
We ran all algorithms independently five times for each task and repeated ten episodes for each evaluation. We reported the mean success rate and standard deviation. The experiments were run on a machine with four NVIDIA A800 GPUs.

\begin{figure*}[htbp]
    \centering
    \includegraphics[width=0.9\textwidth]{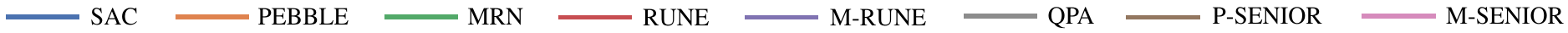} % 调整图例大小
    \vspace{0.5em} % 增加一点垂直间距
    \begin{subfigure}[b]{0.161\textwidth}
        \captionsetup{justification=centering}
        \includegraphics[width=\textwidth]{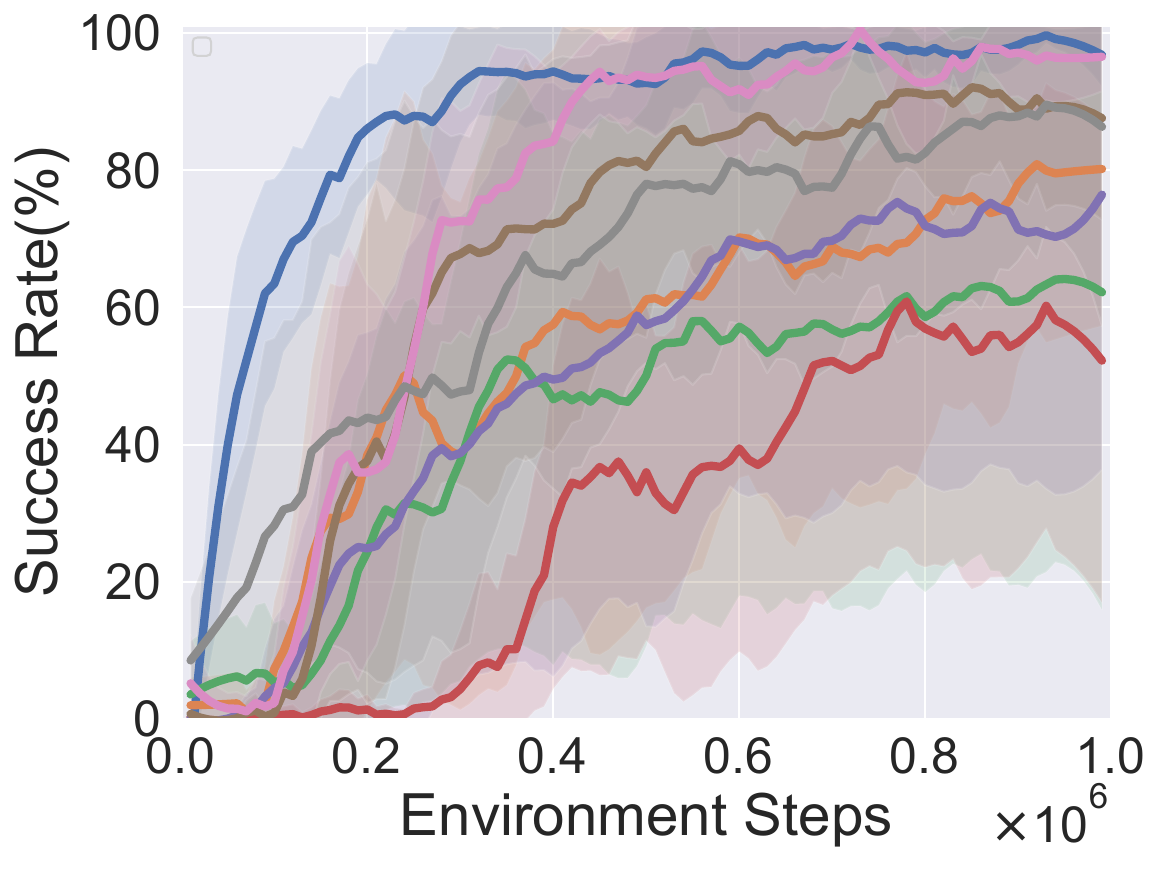}
        \caption{Door Lock \\\quad (feedback=250)}
        \label{experiments_door_lock}
    \end{subfigure}
        \centering
    \begin{subfigure}[b]{0.161\textwidth}
    \captionsetup{justification=centering}
        \includegraphics[width=\textwidth]{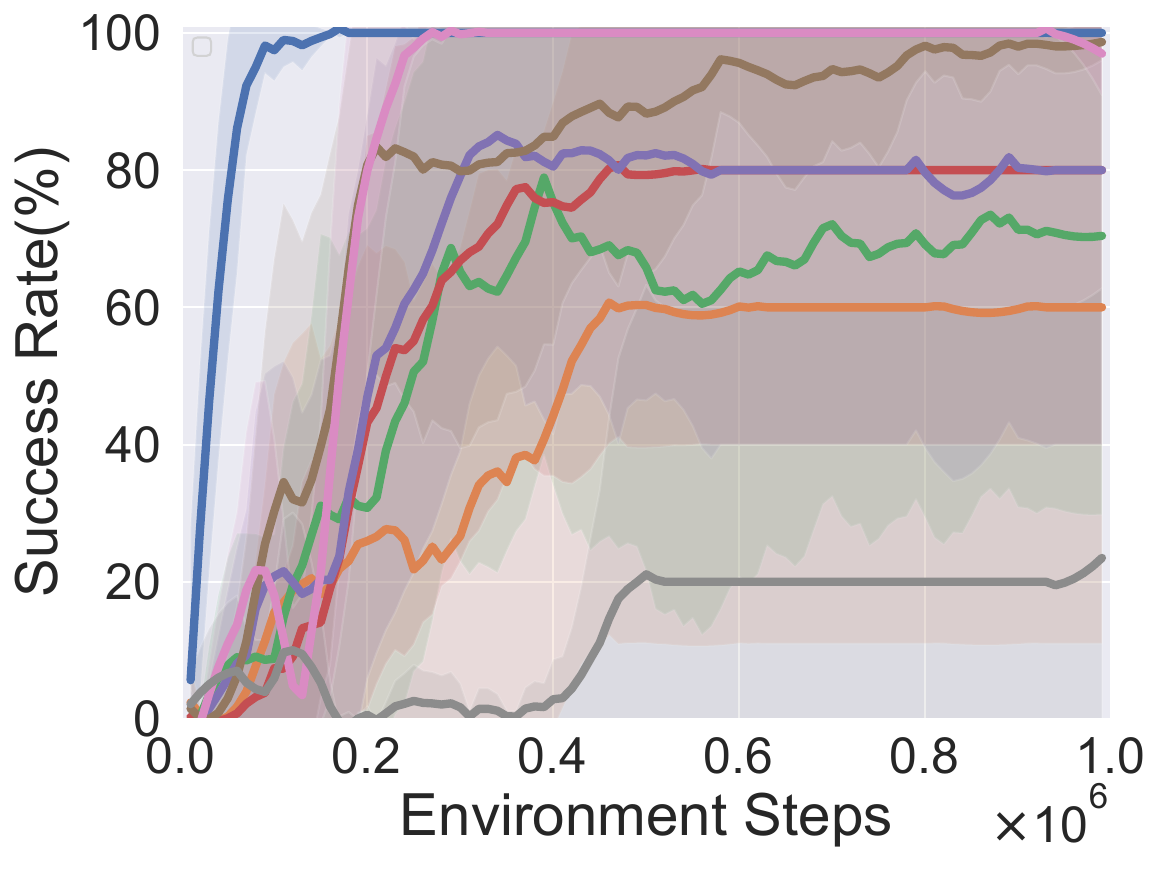}
        \caption{Window Close \\\quad(feedback=250)}
        \label{experiments_window_close}
    \end{subfigure}
    \centering
    \begin{subfigure}[b]{0.161\textwidth}
    \captionsetup{justification=centering}
        \includegraphics[width=\textwidth]{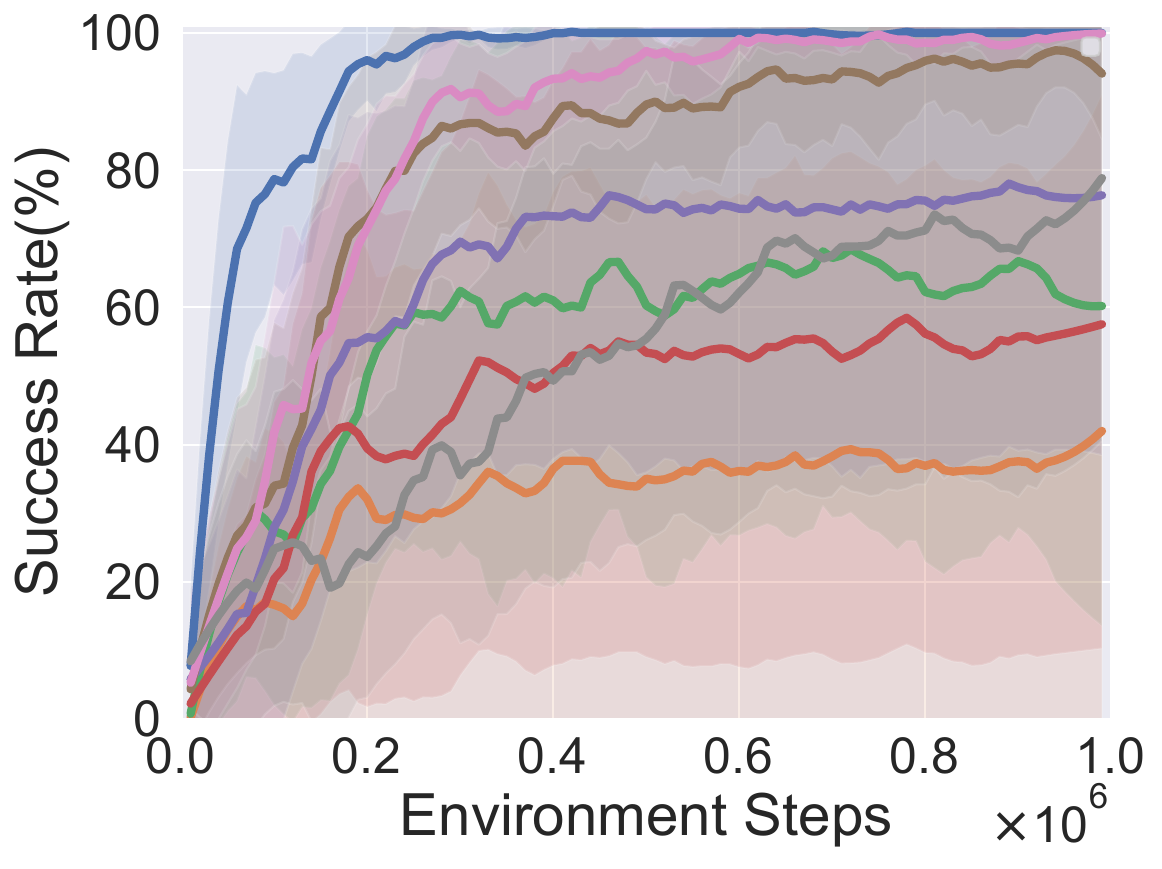}
        \caption{Handle Press \\\quad(feedback=250)}
        \label{experiments_handle_press}
    \end{subfigure}
    \centering
    \begin{subfigure}[b]{0.161\textwidth}
    \captionsetup{justification=centering}
        \includegraphics[width=\textwidth]{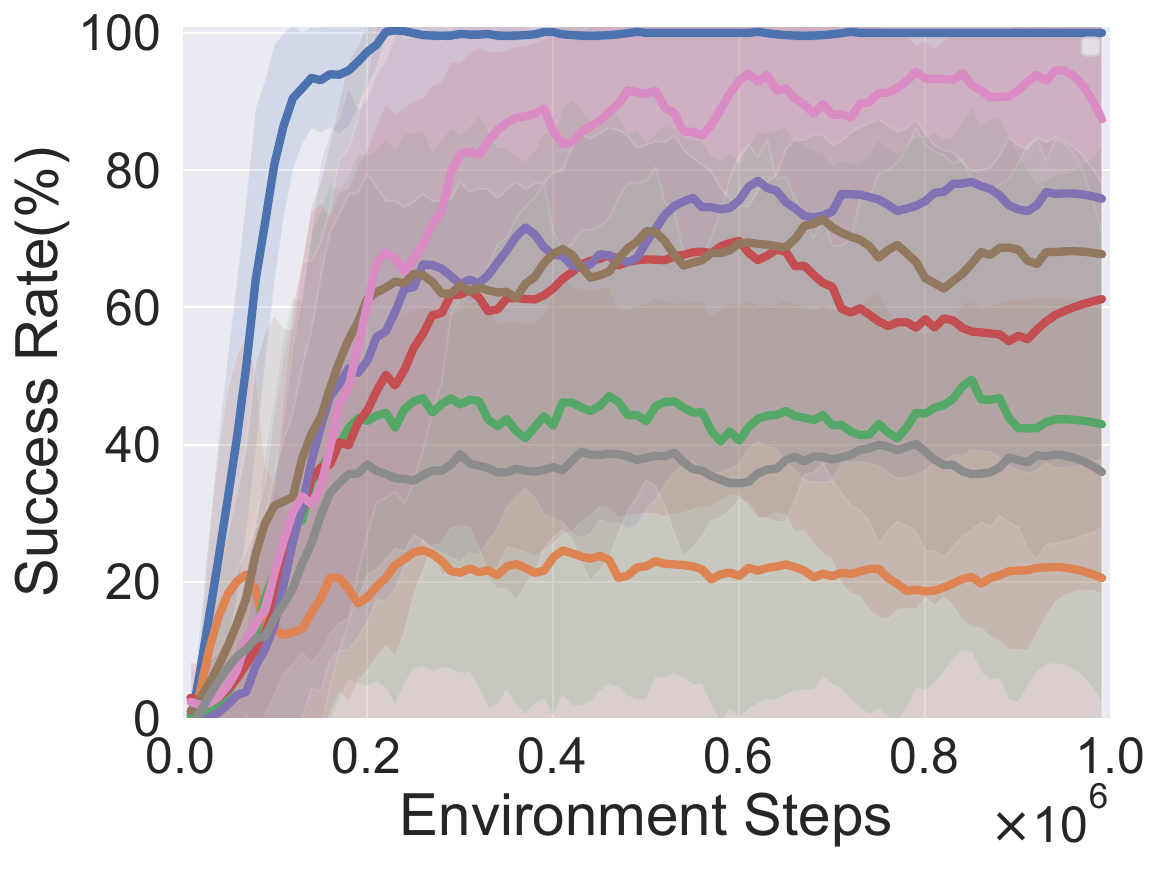}
        \caption{Window Open \\\quad(feedback=250)}
        \label{experiments_window_open}
    \end{subfigure}
    \centering
    \begin{subfigure}[b]{0.161\textwidth}
    \captionsetup{justification=centering}
    \includegraphics[width=\textwidth]{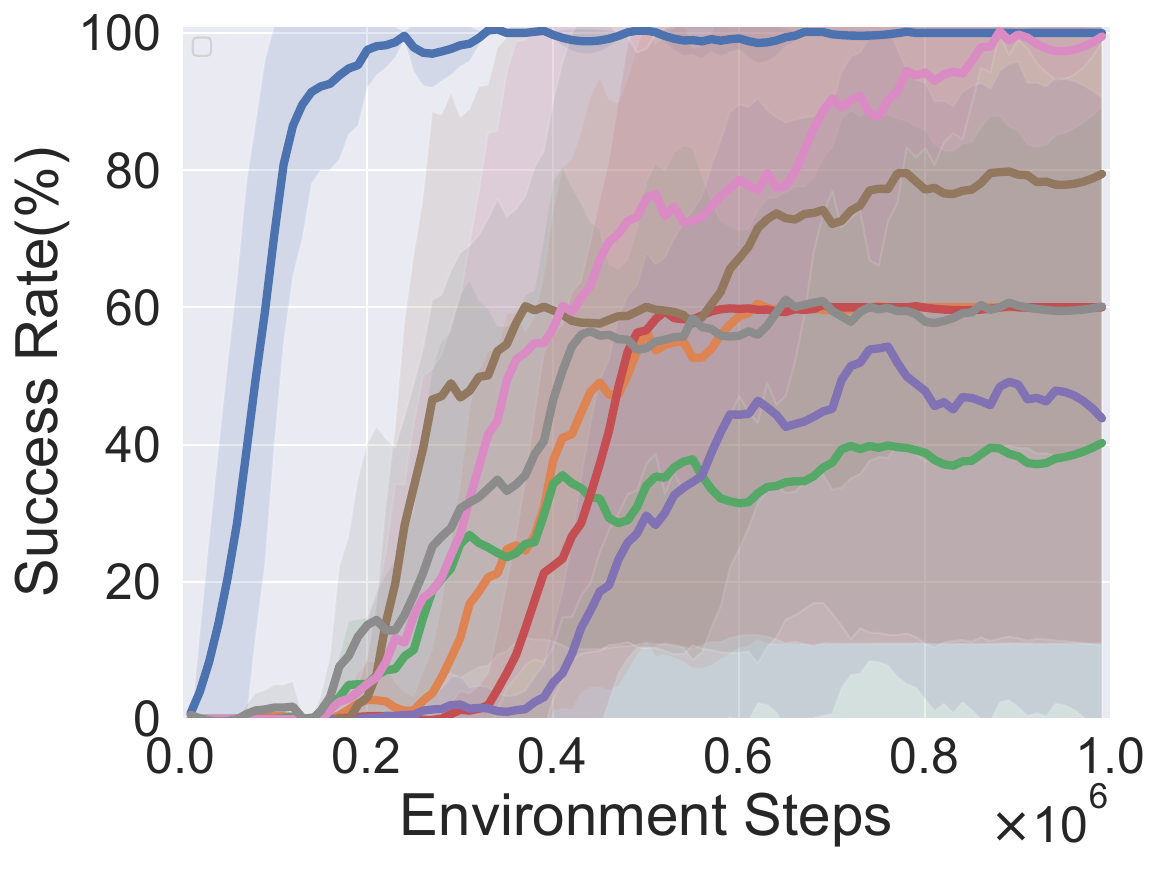}
    \caption{Door Open \\\quad(feedback=1000)}
    \label{experiments_door_open}
    \end{subfigure}
    \centering
    \begin{subfigure}[b]{0.161\textwidth}
    \captionsetup{justification=centering}
    \includegraphics[width=\textwidth]{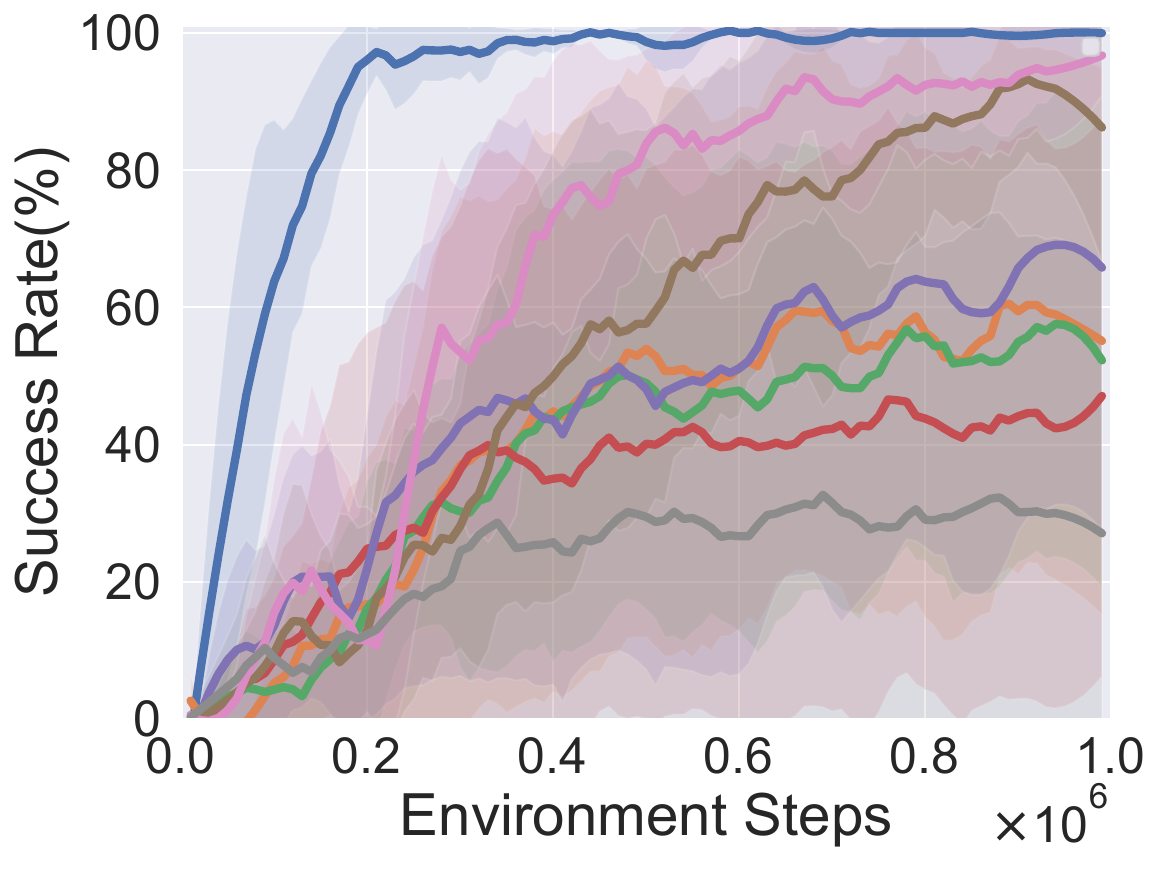}
    \caption{Door Unlock \\\quad(feedback=1000)}
    \label{experiments_door_unlock}
    \end{subfigure}
    \caption{Learning curves on six robotic manipulation tasks as measured by success rate. The solid line and shaded regions represent the mean and standard deviation, respectively, across five runs.}
    \label{fig:experiments}
\end{figure*}

\begin{table*}[htbp]
    \centering
    \renewcommand{\arraystretch}{1.2} % 增加表格行距，提高可读性
    \setlength{\tabcolsep}{5pt} % 调整列间距
    \caption{Comparison of success rates for six tasks at 500K and 1000K steps}
    \begin{tabular}{l rr rr rr rr rr rr | rr}
    \toprule
    \multirow{2}{*}{Method} 
    & \multicolumn{2}{c}{Door Lock} & \multicolumn{2}{c}{Window Close} 
    & \multicolumn{2}{c}{Handle Press} & \multicolumn{2}{c}{Window Open} 
    & \multicolumn{2}{c}{Door Open} & \multicolumn{2}{c|}{Door Unlock} & \multicolumn{2}{c}{Average} \\
    \cmidrule(lr){2-3} \cmidrule(lr){4-5} \cmidrule(lr){6-7}
    \cmidrule(lr){8-9} \cmidrule(lr){10-11} \cmidrule(lr){12-13} \cmidrule(lr){14-15}
    & 500K & 1000K & 500K & 1000K & 500K & 1000K 
    & 500K & 1000K & 500K & 1000K & 500K & 1000K & 500K & 1000K\\
    \midrule
    % SAC & 96\% & 98\% & 100\% & 100\% & 100\% & 100\% & 100\% & 100\% & 100\% & 100\% & 100\% & 100\% & 99\% & 100\% \\
    PEBBLE & 60\% & 82\% & 60\% & 60\% & 36\% & 42\% & 20\% & 20\% & 58\% & 60\% & 56\% & 52\% & 48\% & 53\% \\
    MRN & 41\% & 60\% & 62\% & 66\% & 66\% & 62\% & 52\% & 44\% & 24\% & 38\% & 54\% & 52\% & 50\% & 54\% \\
    RUNE & 34\% & 54\% & 80\% & 80\% & 58\% & 58\% & 70\% & 60\% & 58\% & 60\% & 38\% & 48\% & 56\% & 60\% \\
    M-RUNE & 60\% & 76\% & 80\% & 80\% & 76\% & 76\% & 62\% & 78\% & 26\% & 42\% & 48\% & 66\% & 59\% & 70\% \\
    QPA & 76\% & 90\% & 20\% & 26\% & 56\% & 80\% & 40\% & 36\% & 48\% & 60\% & 30\% & 28\% & 45\% & 53\% \\
    \addlinespace[0.1em] \midrule
    \textbf{P-SENIOR(Ours)} & \textbf{86\%} & \textbf{92\%} & \textbf{88\%} & \textbf{98\%} 
    & \textbf{90\%} & \textbf{94\%} & \textbf{68\%} & \textbf{68\%} 
    & \textbf{60\%} & \textbf{80\%} & \textbf{64\%} & \textbf{84\%}
    & \textbf{76\%} & \textbf{86\%} \\
    \textbf{M-SENIOR(Ours)} & \textbf{92\%} & \textbf{98\%} & \textbf{100\%} & \textbf{100\%} 
    & \textbf{98\%} & \textbf{100\%} & \textbf{88\%} & \textbf{90\%} 
    & \textbf{72\%} & \textbf{100\%} & \textbf{76\%} & \textbf{96\%}
    & \textbf{88\%} & \textbf{97\%} \\
    \bottomrule
    \end{tabular}
    \label{tab:sample_efficiency}
\end{table*}

\begin{figure}[htbp!]
    \centering
    \begin{subfigure}[b]{0.235\textwidth}
        \includegraphics[width=\textwidth]{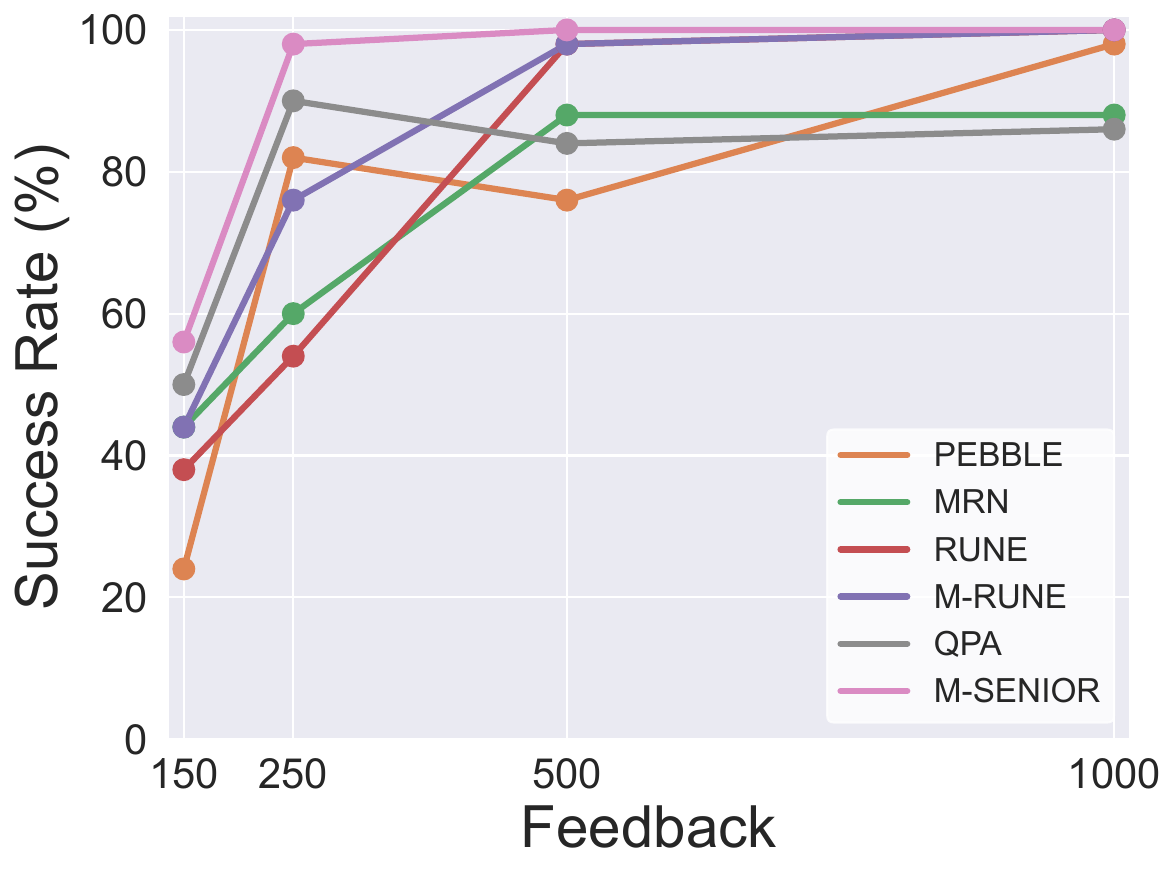}
        \caption{Door Lock}
    \end{subfigure}
    \begin{subfigure}[b]{0.235\textwidth}
        \includegraphics[width=\textwidth]{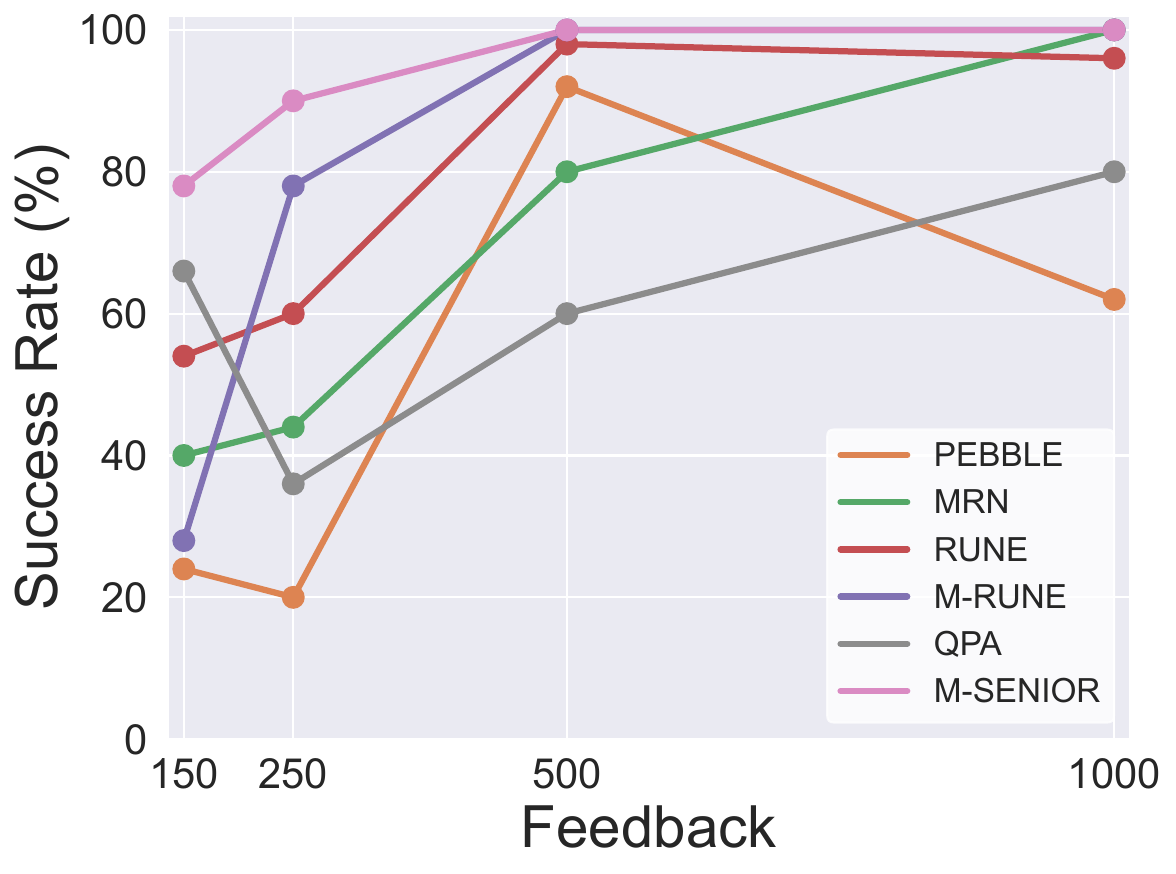}
        \caption{Window Open}
    \end{subfigure}
    \caption{Comparison of final success rates for different feedback budgets on Door Lock and Window Open.}
    \label{fig:feedback_efficiency}
\end{figure}
\begin{figure}[htbp!]
    \centering
    \begin{subfigure}[b]{0.4\textwidth}
        \includegraphics[width=\textwidth]{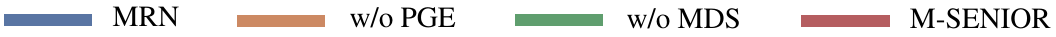}
    \end{subfigure}
    \begin{subfigure}[b]{0.42\textwidth}
        \includegraphics[width=\textwidth]{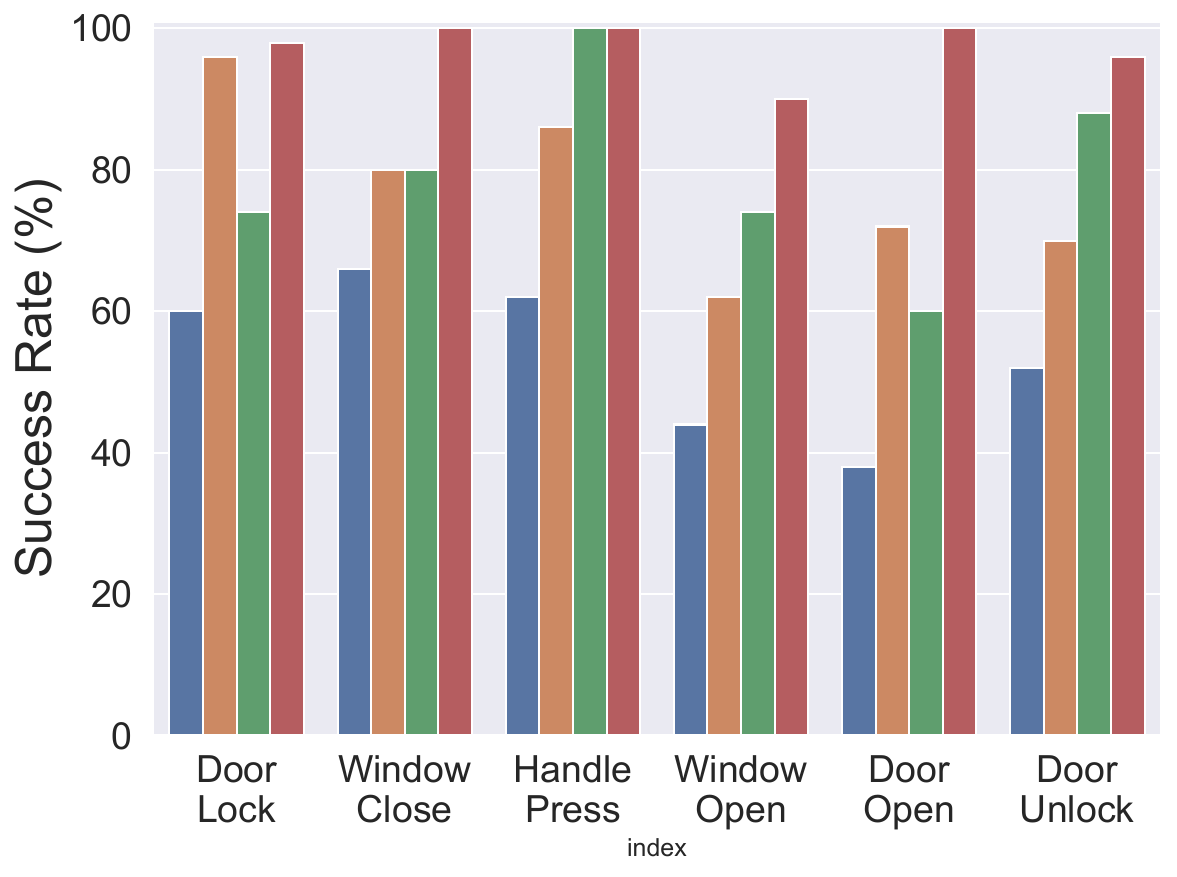}
    \end{subfigure}
    \caption{Ablation study on six tasks as measured by success rate. The final result ran independently five times.}
    \label{fig:ablation}
\end{figure}

\subsection{Results}
Fig.~\ref{fig:experiments} illustrates task success rate curves during training. With less feedback budget for all tasks, our methods achieved the best performance and convergence speed compared to others. As shown in Table \ref{tab:sample_efficiency}, M-SENIOR has a final average success rate of 97\% on the six tasks and P-SENIOR has 86\%. M-RUNE, RUNE and QPA only achieve 70\%, 60\% and 53\% respectively. 
M-SENIOR reaches averagely 88\% success rate at 500K steps, which far exceeds other methods at 500K, even until 1000K steps. 
Especially in Door Open task, high-quality feedback provided by SENIOR is critical for sustained performance improvement, and M-SENIOR ultimately outperforms other methods by 40\%. 
Without bi-level optimization, P-SENIOR may not perform as well as M-SENIOR (In Window
Open, the success rate (68\%) is lower than M-RUNE (78\%)), but in most cases, P-SENIOR
still outperforms the baselines, notably in improving the average performance by 33\%
compared to PEBBLE. This further verifies the effectiveness of our method.

To verify the feedback-efficiency, we studied the final performance of different methods under 150, 250, 500 and 1000 feedback budgets in Door Lock and Window Open tasks.
As shown in Fig. \ref{fig:feedback_efficiency}, it is evident that our method achieves the highest success rate in all settings and could increase at least $4 \times$ feedback-efficiency. For example, M-SENIOR reaches nearly 100\% success rate with only 250 feedback on Door Lock, but other methods require 1000 feedback or more. 
Moreover, the performance of M-SENIOR improves continuously when feedback increases, which is not always the case for other methods. This is because SENIOR can consistently select helpful feedback, whereas other query selection schemes may introduce noise feedback and impair reward learning during training.

\subsection{Ablation Experiment}
We also performed ablation experiments on the six tasks to measure the effect of each component (MDS and PGE) in SENIOR. Fig.~\ref{fig:ablation} shows the results under the same settings as above. 
We can see that w/o PGE and w/o MDS all performed better than MRN, and the combination of both has the highest learning efficiency. This shows that each component of our method is helpful in finding out valuable samples (high-quality feedback labels or exploration states) during training. In some cases, one component may greatly improve the performance, such as MDS mainly guided the policy learning in Door Lock, PGE performed well in Handle Press. Generally, the synergy between the two components is the key to the success of our method.

\begin{figure}[htbp!]
    \centering
    \begin{subfigure}[b]{0.235\textwidth}
        \includegraphics[width=\textwidth]{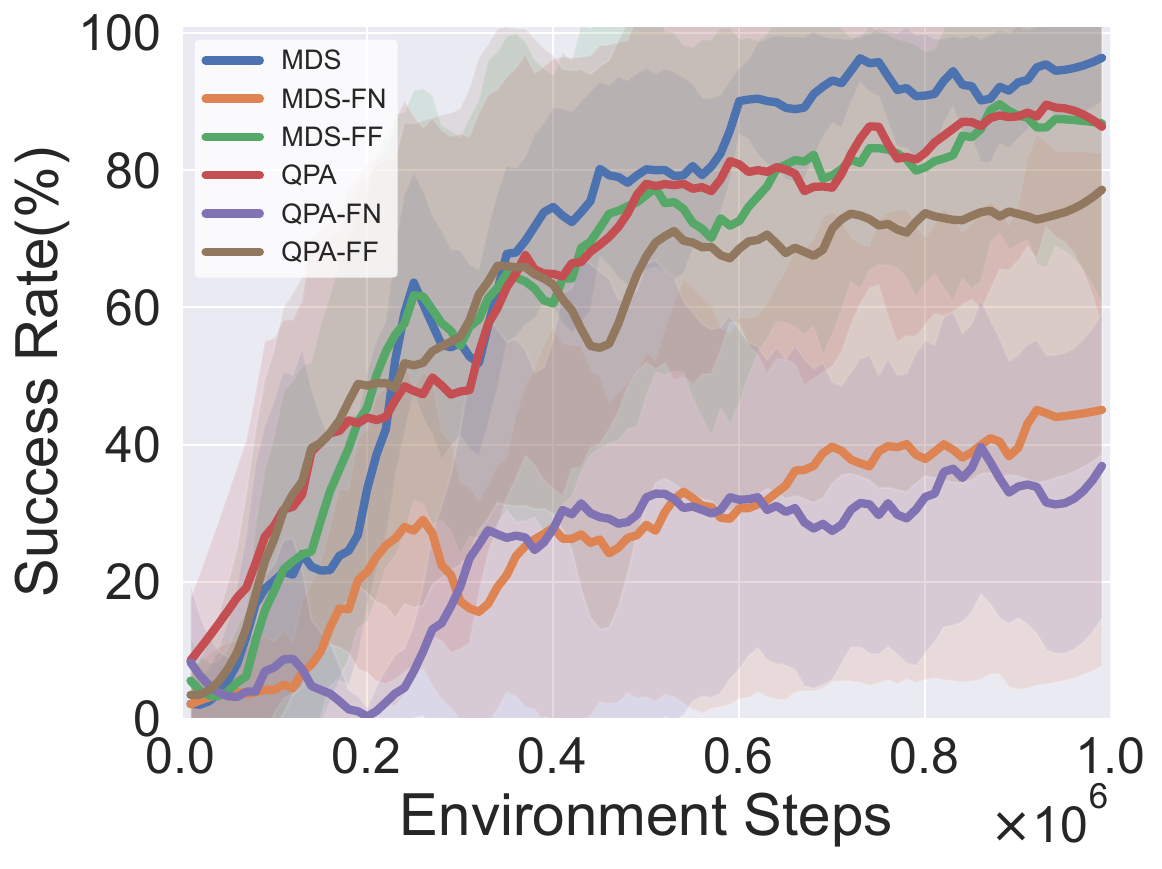}
        \caption{Door Lock}
    \end{subfigure}
    \begin{subfigure}[b]{0.235\textwidth}
        \includegraphics[width=\textwidth]{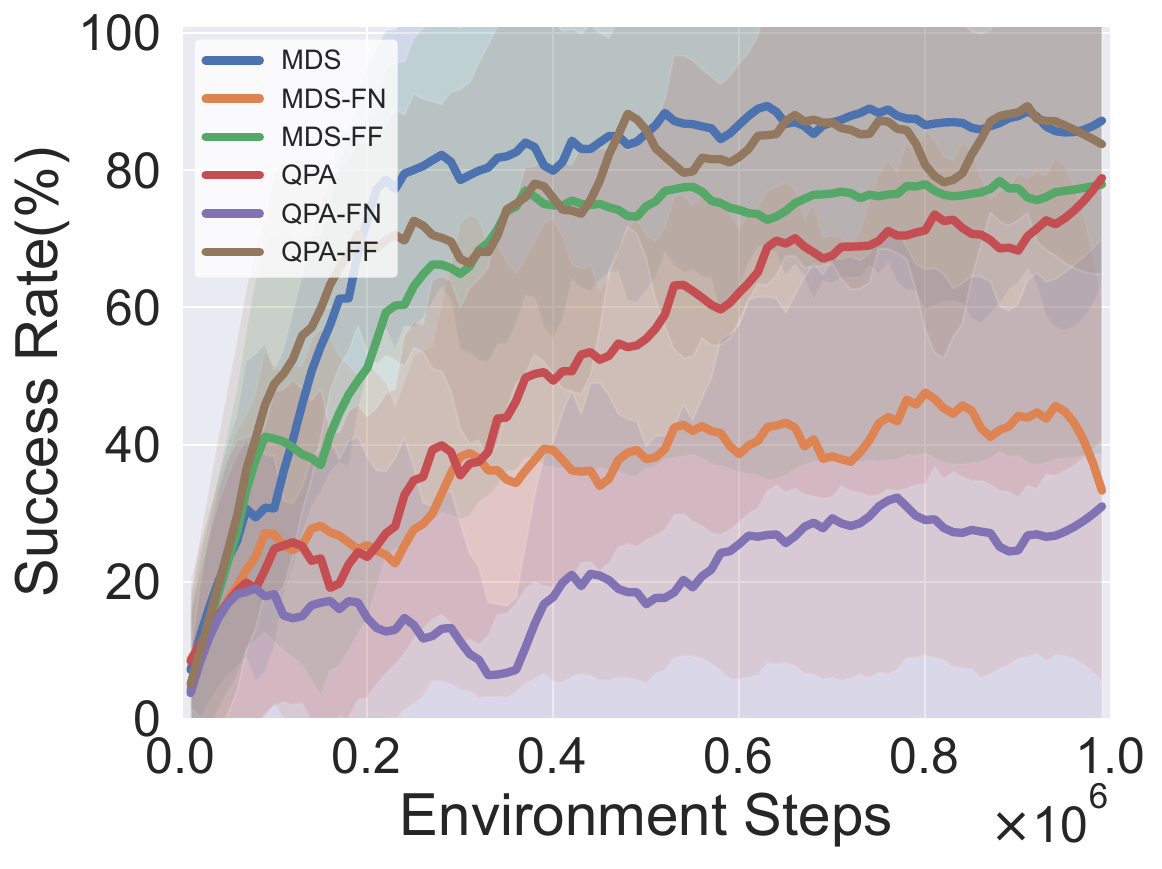}
        \caption{Handle Press}
    \end{subfigure}
    \caption{Influence of feedback quality on Door Lock and Handle Press tasks. }
    \label{fig:misalignment}
\end{figure}

\textbf{Influence of Feedback Quality.}
To further show the informative and meaningful segments are important in PbRL, we study the effect of feedback quality on reward and policy learning. We compared the performance of MDS with QPA in Door Lock and Handle Press tasks. During training, we added \textbf{\underline{F}}eedback \textbf{\underline{N}}oise (-FN, 10\% wrong feedback) and \textbf{\underline{F}}eedback \textbf{\underline{F}}ilter (-FF, neglect queries with low environment rewards) to MDS and QPA respectively.  As shown in Fig. \ref{fig:misalignment}, the performance of MDS-FN and QPA-FN decreases dramatically. This shows the noise feedback seriously affects the performance, which on the other hand means that our MDS has the ability to achieve high-quality feedback and align the query with policy as QPA. For MDS-FN and QPA-FN, we can see that the success rate of MDS-FN becomes lower than that of MDS in both tasks, but QPA-FN even has a higher success rate than QPA in Handle Press. This shows that our method could effectively and stably use the low environment reward samples to improve the performance.

\begin{figure}[htbp]
    \centering
    \includegraphics[width=0.46\textwidth]{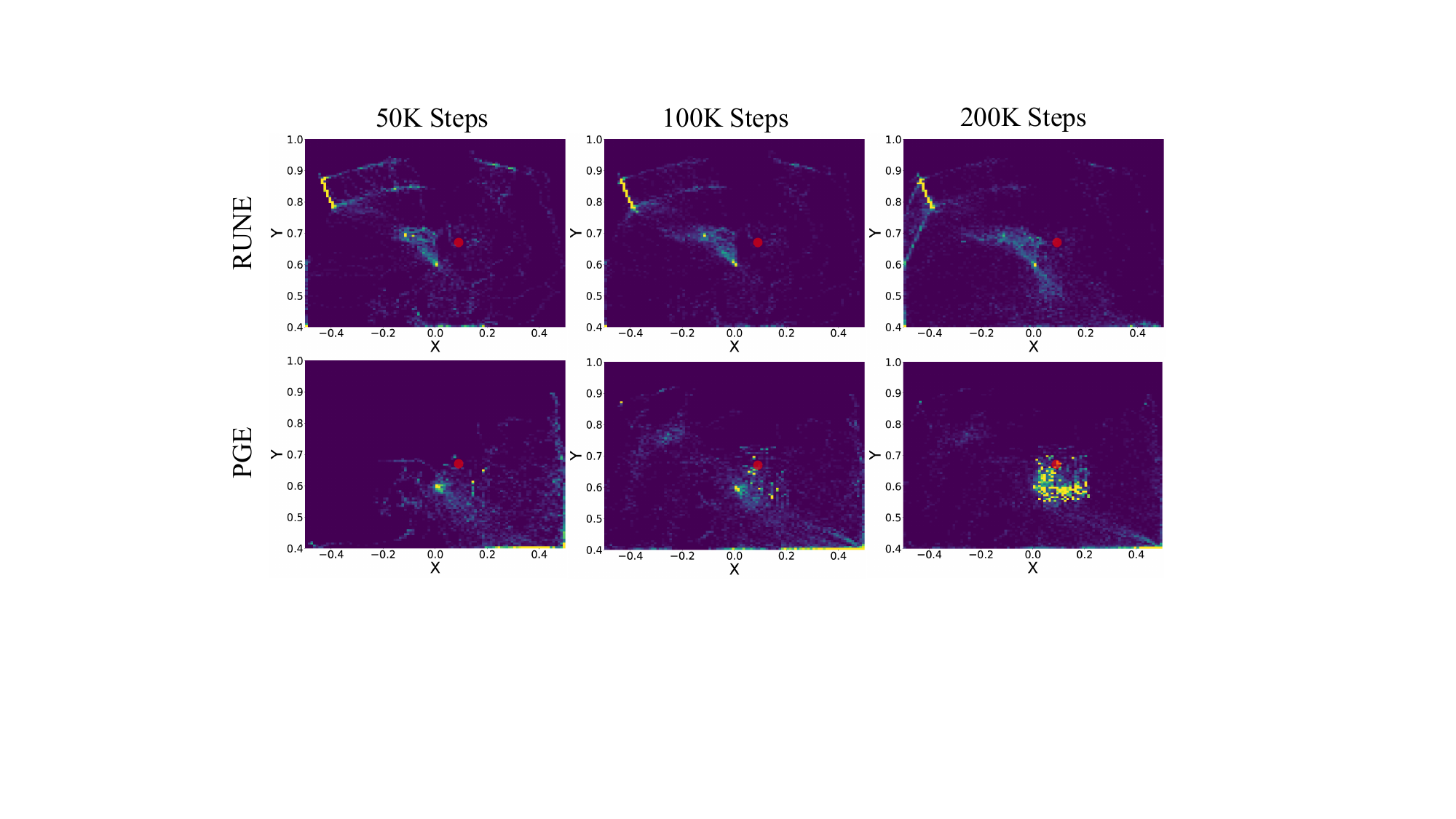}
    \caption{Visualization of state visitation distribution in the Door Lock task environment during training. The red dot marks the lock position and bright yellow dots indicate the explored states. }
    \label{fig:exploration}
\end{figure}
\textbf{Preference guidance for Exploration.}
To better analyze our exploration mechanism, we visualized the state visitation distribution of the robot in the Door Lock task during training. 
We show the performance of PGE and RUNE respectively in Fig. \ref{fig:exploration}. Bright yellow dots indicate the explored states and the red indicates the lock position. It can be seen clearly that PGE becomes focusing on exploring areas of the lock preferred by humans in 200K steps, but RUNE has no apparent exploration around the lock. The reason is that RUNE prioritizes exploring areas with high reward uncertainty and does not supply more task-relevant information. Our PGE could better converge the exploration towards the task goal under the guidance of human preference. 
\subsection{Deployment on a Physical Robot}

\begin{figure}[htbp!]
    \centering
    \begin{subfigure}[b]{0.5\textwidth}
        \centering
        \captionsetup{skip=1.0pt}
        \begin{subfigure}[b]{0.18\textwidth}
        \centering
            \includegraphics[width=\textwidth]{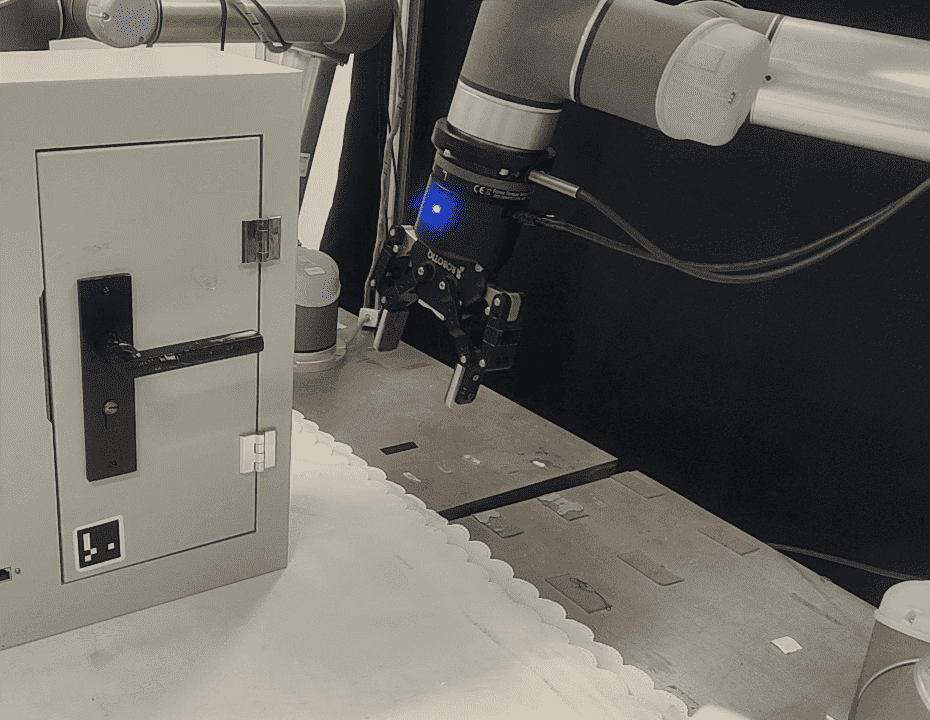}
        \end{subfigure}
        \hspace{-0.5em} % 调整水平间距
        \begin{subfigure}[b]{0.18\textwidth}
            \includegraphics[width=\textwidth]{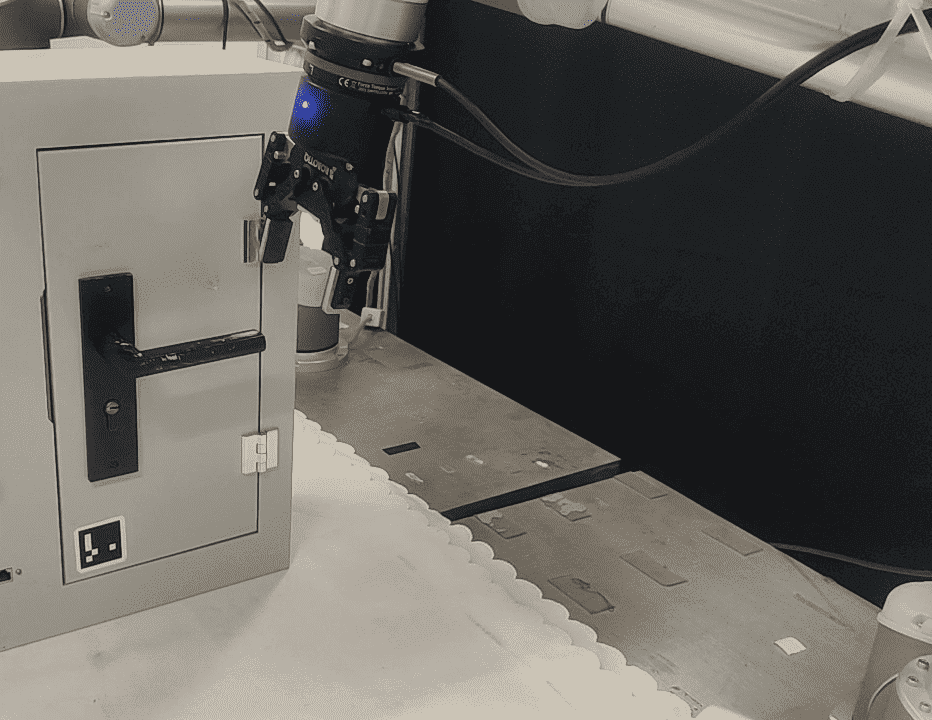}
        \end{subfigure}
        \hspace{-0.5em} % 调整水平间距
        \begin{subfigure}[b]{0.18\textwidth}
            \includegraphics[width=\textwidth]{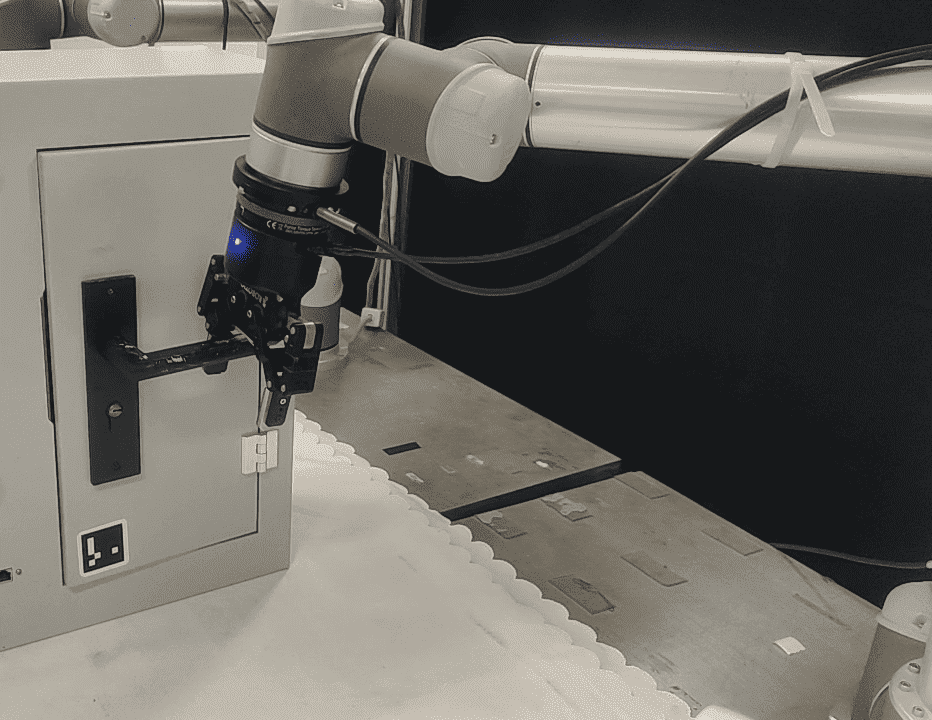}
        \end{subfigure}
        \hspace{-0.5em} % 调整水平间距
        \begin{subfigure}[b]{0.18\textwidth}
            \includegraphics[width=\textwidth]{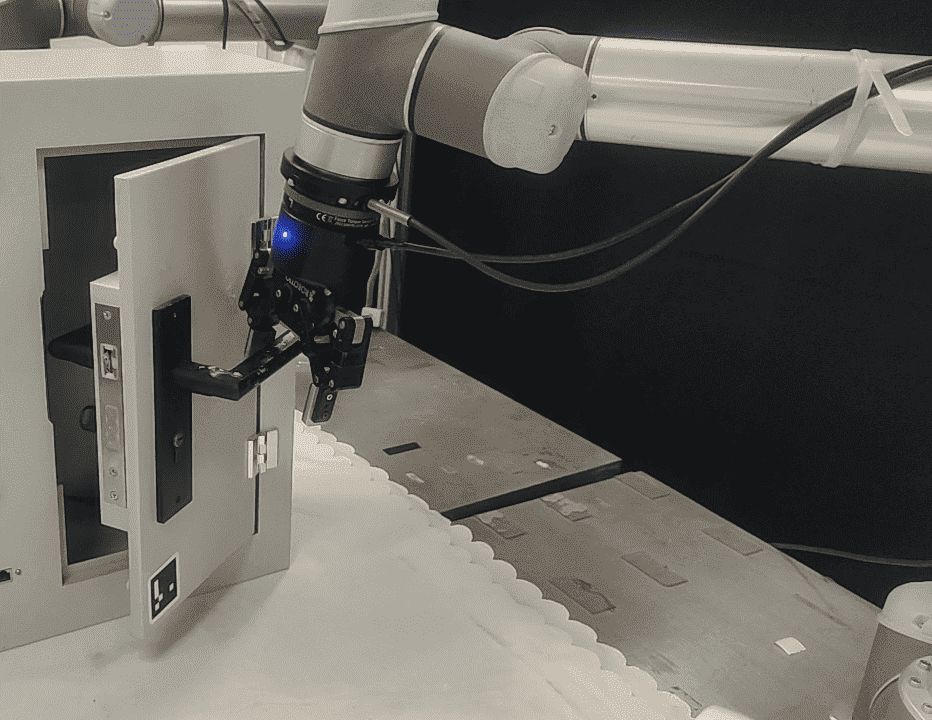}
        \end{subfigure}
        \hspace{-0.5em} % 调整水平间距
        \begin{subfigure}[b]{0.18\textwidth}
            \includegraphics[width=\textwidth]{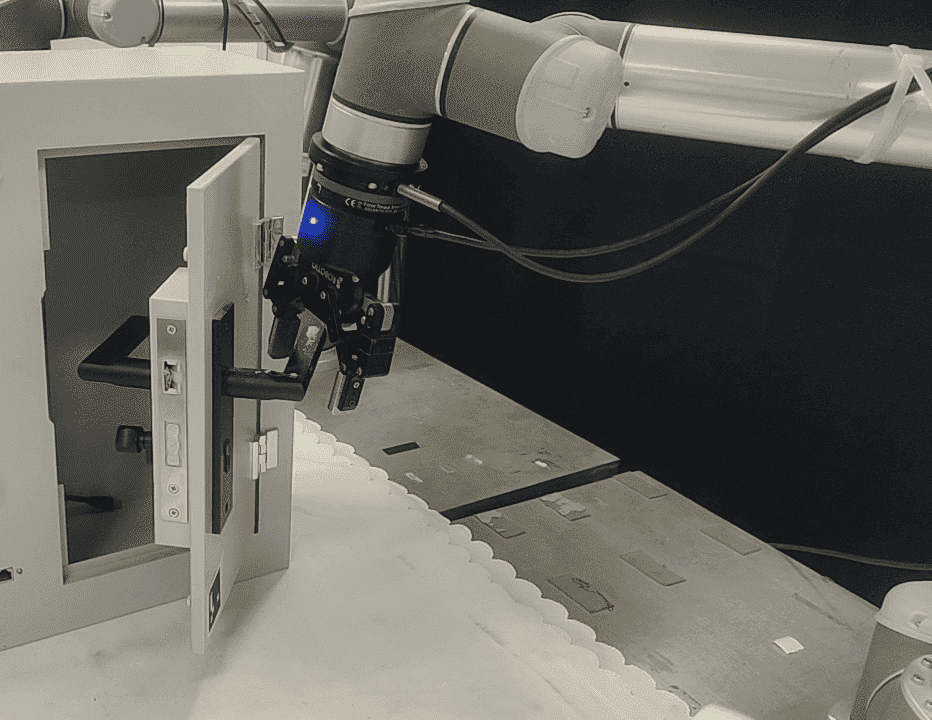}
        \end{subfigure}
        \caption{}
    \end{subfigure}
    
    \vspace{0.2em}
    \centering
    \begin{subfigure}[b]{0.5\textwidth}
        \centering
        \captionsetup{skip=1.0pt}
        \begin{subfigure}[b]{0.18\textwidth}
        \centering
            \includegraphics[width=\textwidth]{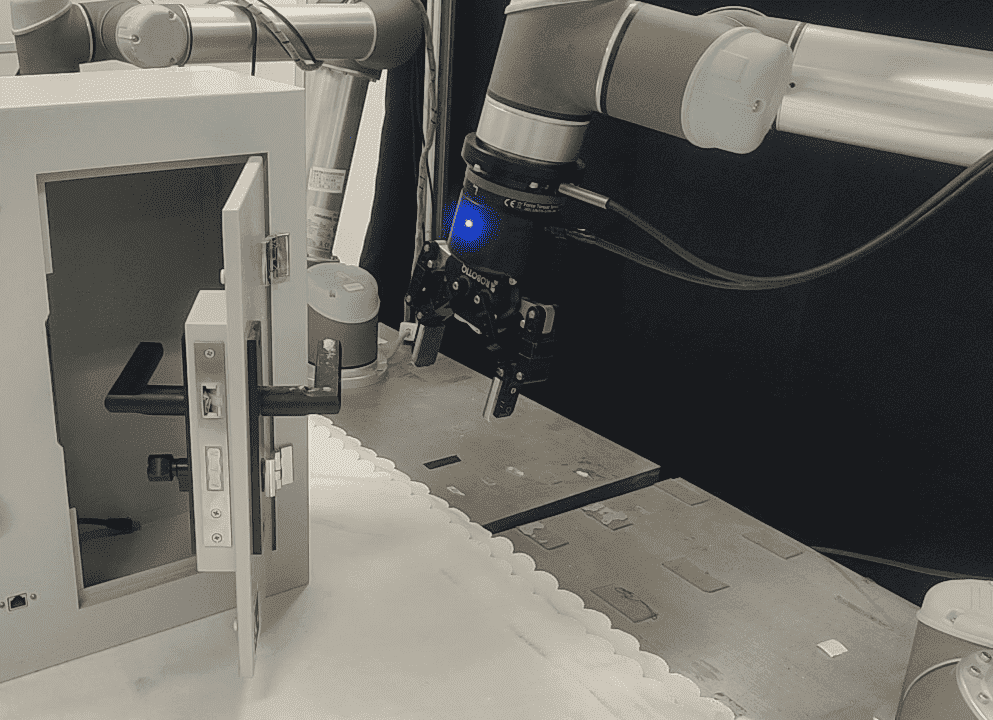}
        \end{subfigure}
        \hspace{-0.5em} % 调整水平间距
        \begin{subfigure}[b]{0.18\textwidth}
            \centering
            \includegraphics[width=\textwidth]{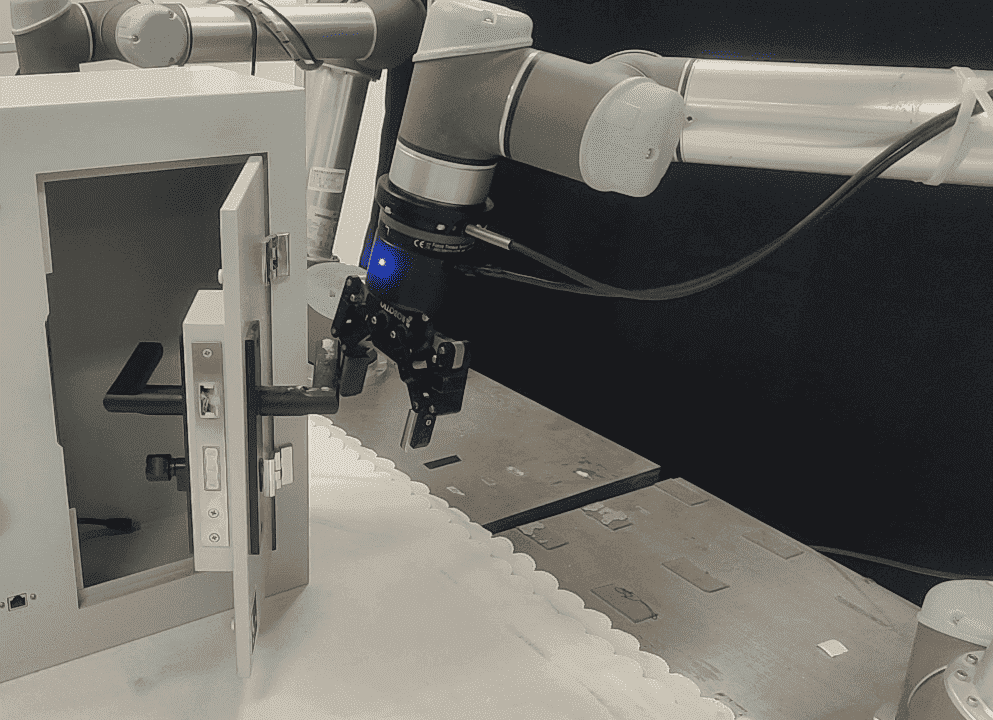}
        \end{subfigure}
        \hspace{-0.5em} % 调整水平间距
        \begin{subfigure}[b]{0.18\textwidth}
        \centering
            \includegraphics[width=\textwidth]{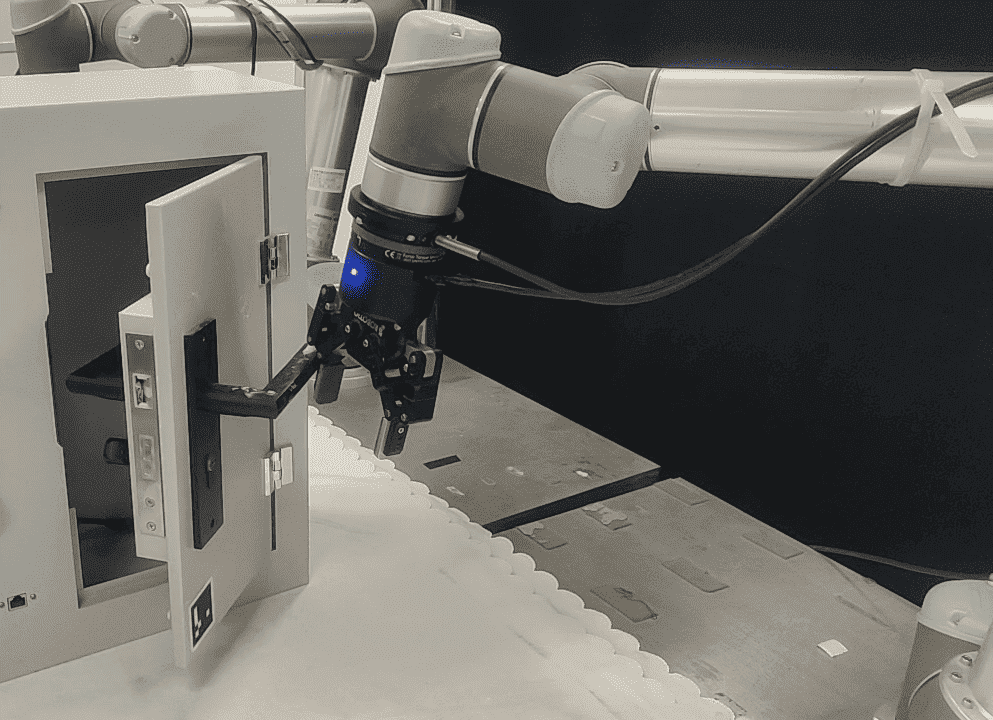}
        \end{subfigure}
        \hspace{-0.5em} % 调整水平间距
        \begin{subfigure}[b]{0.18\textwidth}
            \includegraphics[width=\textwidth]{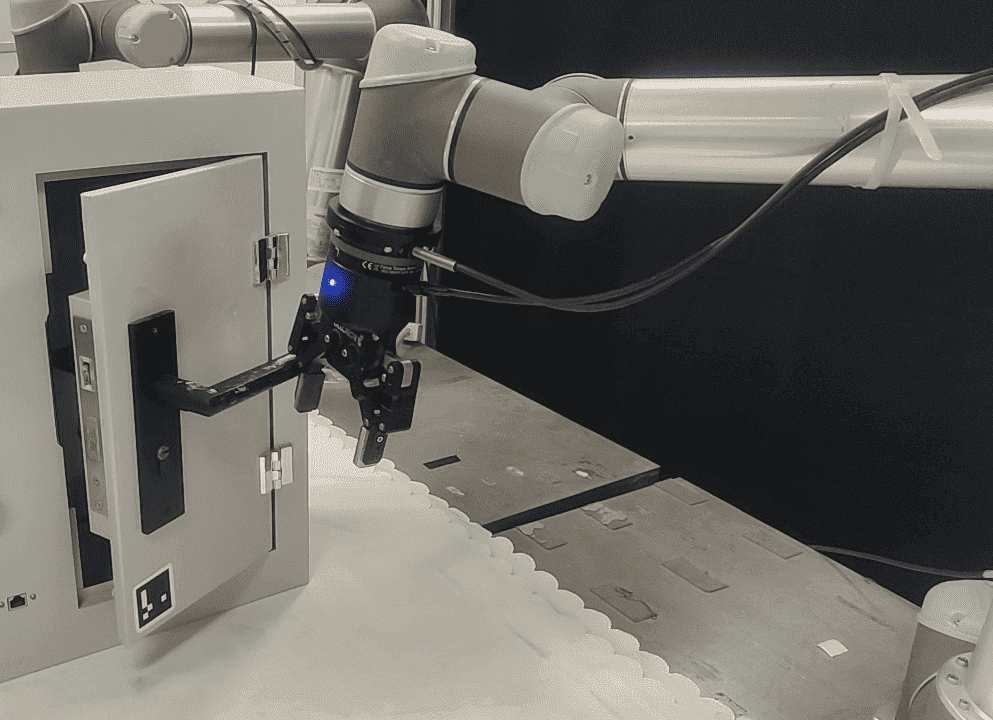}
        \end{subfigure}
        \hspace{-0.5em} % 调整水平间距
        \begin{subfigure}[b]{0.18\textwidth}
        \centering
            \includegraphics[width=\textwidth]{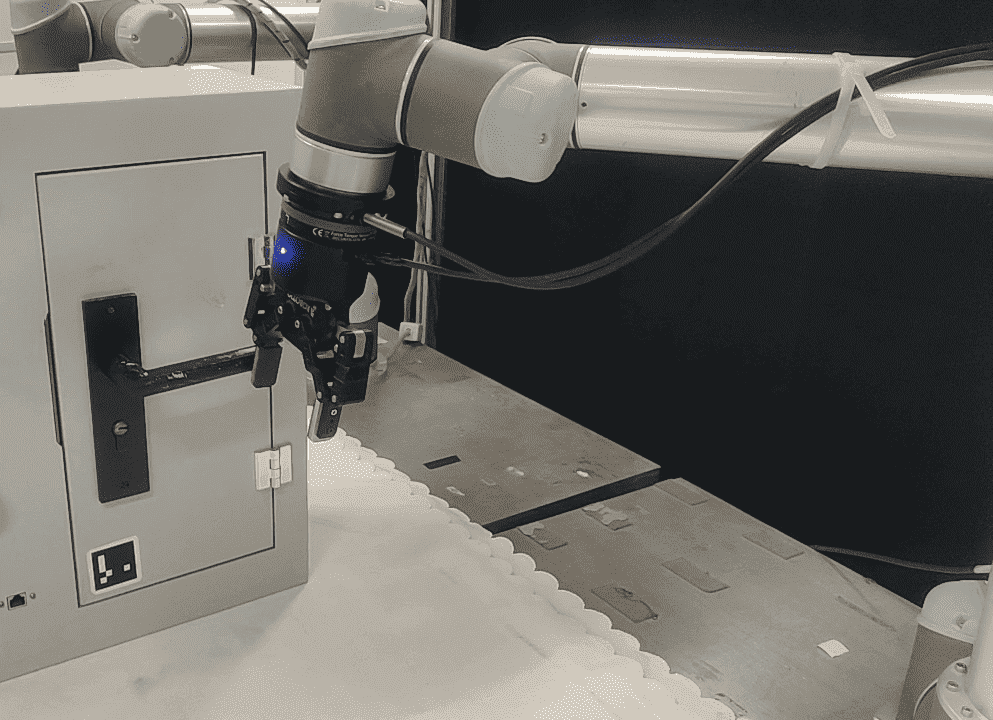}
        \end{subfigure}
        \caption{}
    \end{subfigure}
    
    \vspace{0.2em}
    \centering
    \begin{subfigure}[b]{0.5\textwidth}
        \centering
        \captionsetup{skip=1.0pt}
        \begin{subfigure}[b]{0.18\textwidth}
        \centering
            \includegraphics[width=\textwidth]{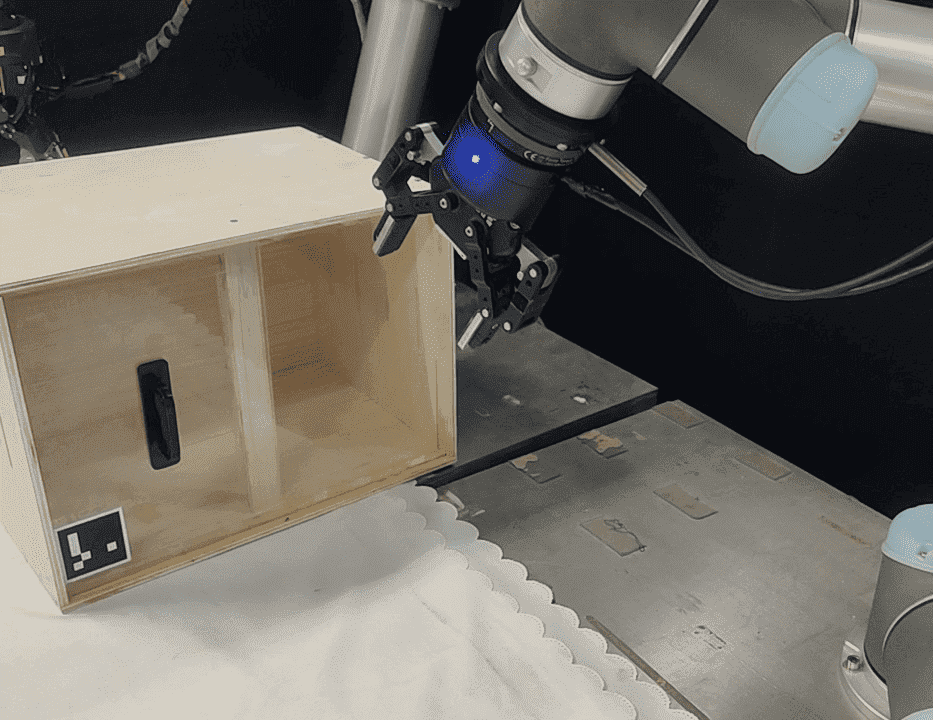}
        \end{subfigure}
        \hspace{-0.5em} % 调整水平间距
        \begin{subfigure}[b]{0.18\textwidth}
            \includegraphics[width=\textwidth]{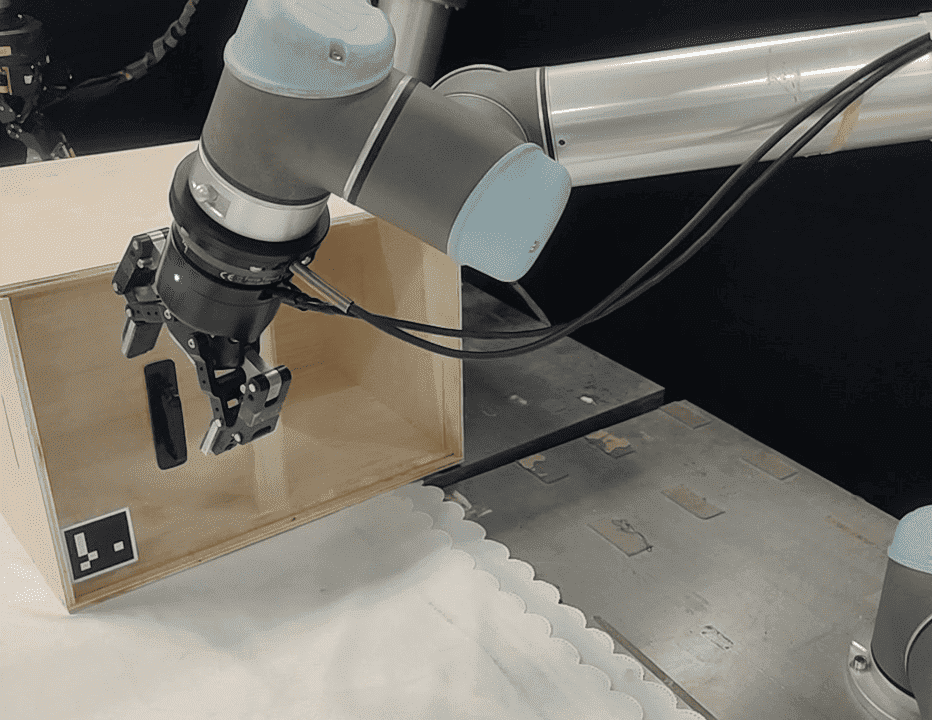}
        \end{subfigure}
        \hspace{-0.5em} % 调整水平间距
        \begin{subfigure}[b]{0.18\textwidth}
            \includegraphics[width=\textwidth]{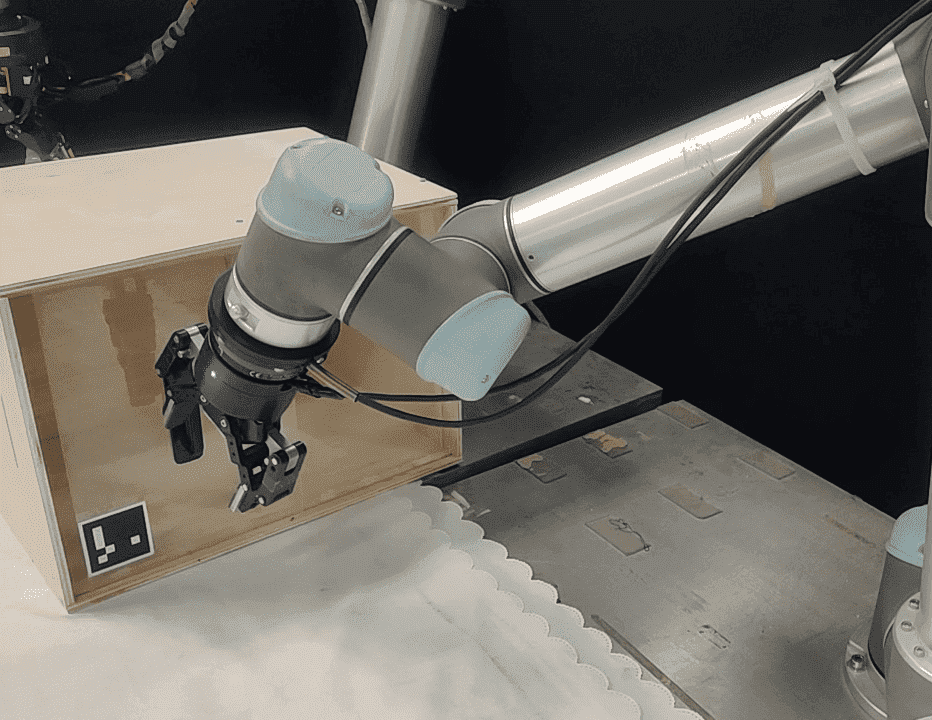}
        \end{subfigure}
        \hspace{-0.5em} % 调整水平间距
        \begin{subfigure}[b]{0.18\textwidth}
            \includegraphics[width=\textwidth]{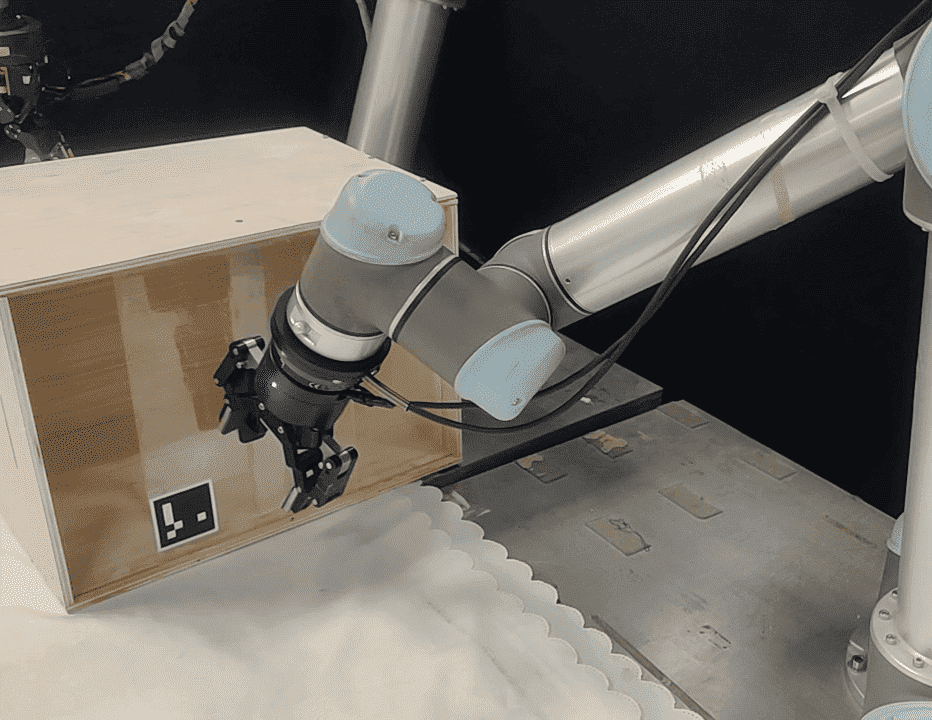}
        \end{subfigure}
        \hspace{-0.5em} % 调整水平间距
        \begin{subfigure}[b]{0.18\textwidth}
            \includegraphics[width=\textwidth]{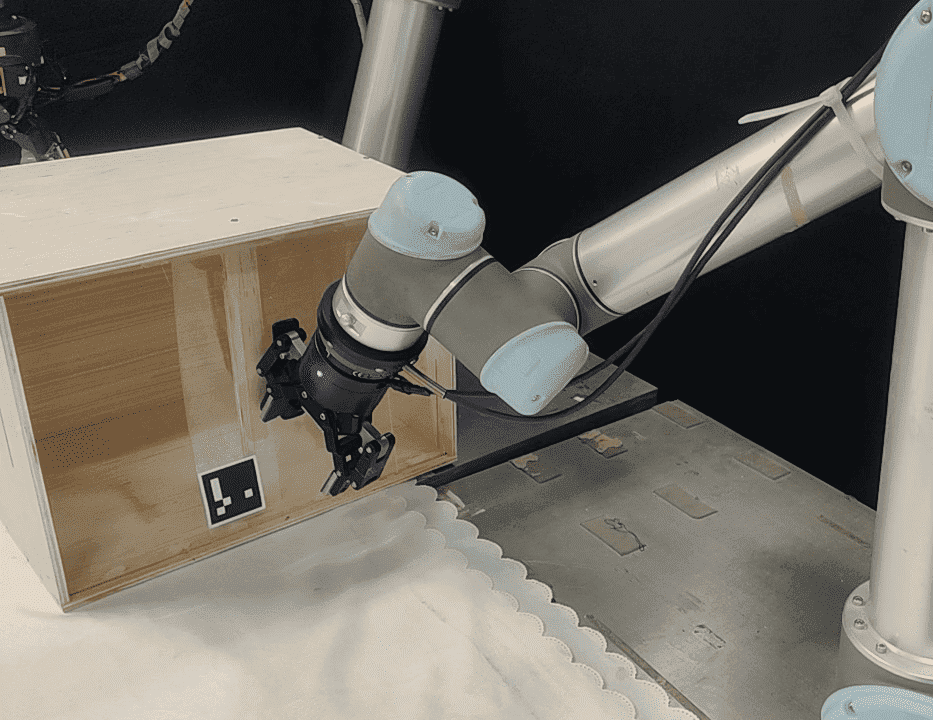}
        \end{subfigure}
        \caption{}
    \end{subfigure}
    
    \vspace{0.2em}
    \centering
    \begin{subfigure}[b]{0.5\textwidth}
        \centering
        \captionsetup{skip=1.0pt}
        \begin{subfigure}[b]{0.18\textwidth}
        \centering
            \includegraphics[width=\textwidth]{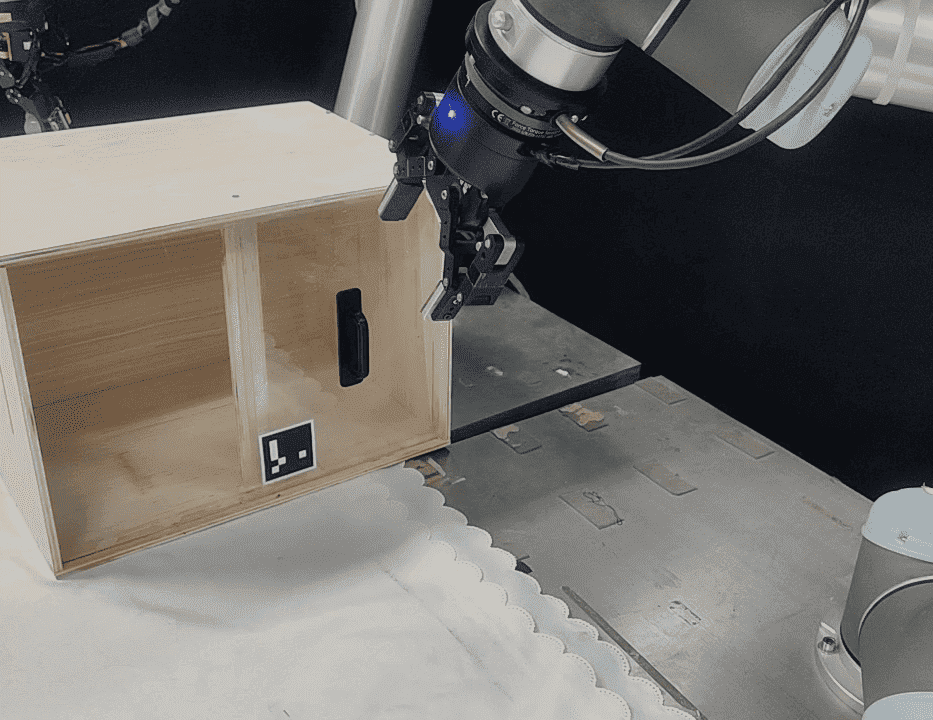}
        \end{subfigure}
        \hspace{-0.5em} % 调整水平间距
        \begin{subfigure}[b]{0.18\textwidth}
            \includegraphics[width=\textwidth]{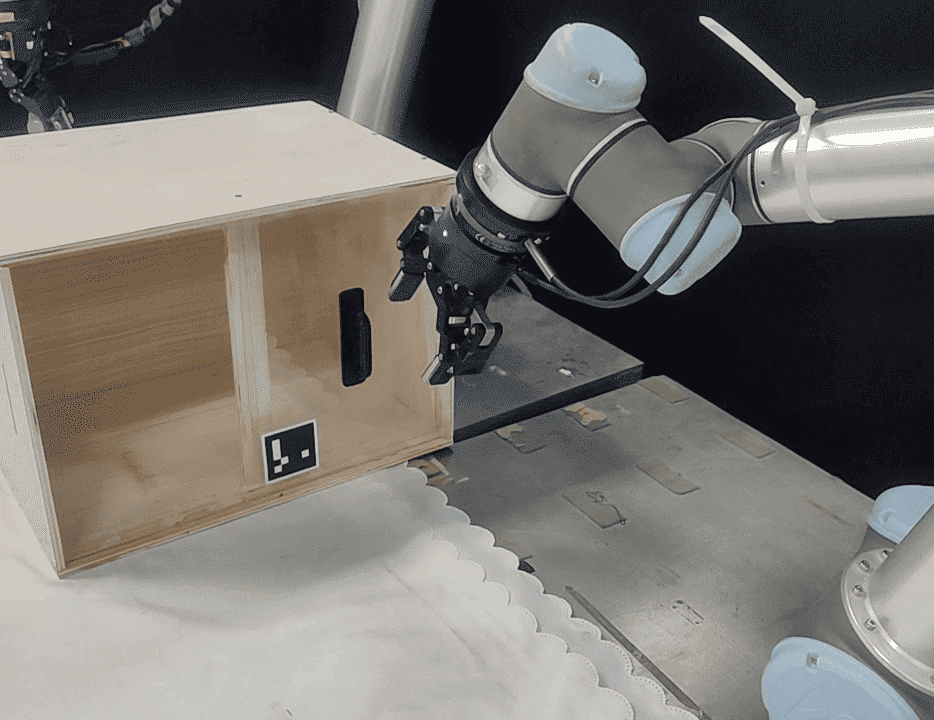}
        \end{subfigure}
        \hspace{-0.5em} % 调整水平间距
        \begin{subfigure}[b]{0.18\textwidth}
            \includegraphics[width=\textwidth]{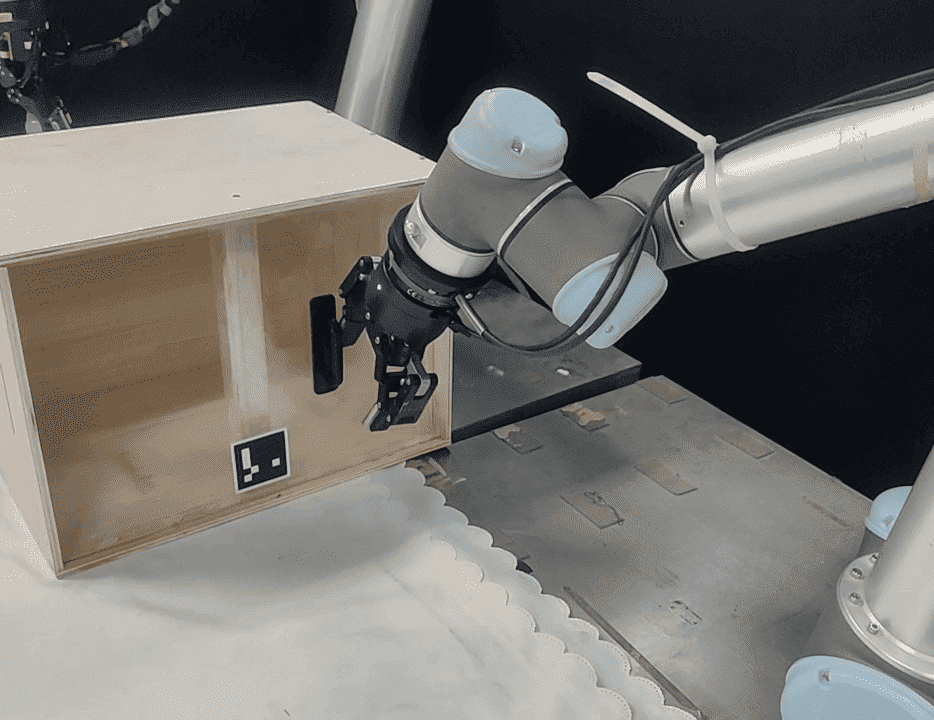}
        \end{subfigure}
        \hspace{-0.5em} % 调整水平间距
        \begin{subfigure}[b]{0.18\textwidth}
            \includegraphics[width=\textwidth]{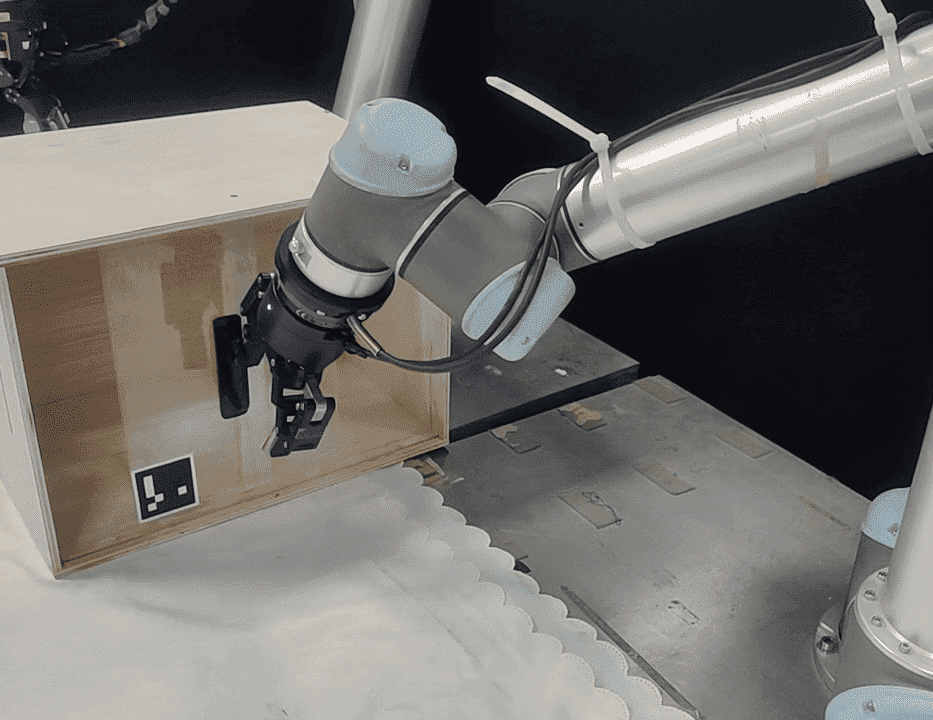}
        \end{subfigure}
        \hspace{-0.5em} % 调整水平间距
        \begin{subfigure}[b]{0.18\textwidth}
            \includegraphics[width=\textwidth]{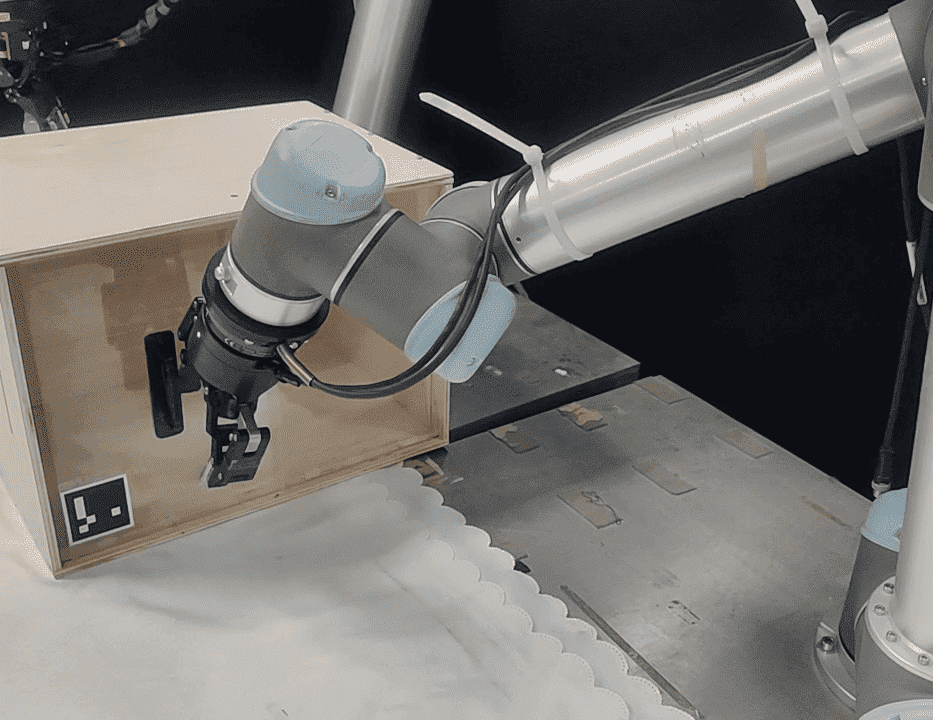}
        \end{subfigure}
        \caption{}
    \end{subfigure}
    
    \caption{Execution sequence of policy on real-world UR5 robotic arm for Door Open (a), Door Close (b), Box Open (c) and Box Close (d) tasks. }
    \label{fig:real}
\end{figure}

We deployed the policies on the Door Open, Door Close, Box Open and Box Close in the real world, as shown in Fig.~\ref{fig:real}. To maintain consistency between the simulation and the real setup, we first modified the simulation environment of these tasks in Meta-World based on the real environment and retrained the policies. Since the policy output is the increment of the end-effector of the robotic arm, the policy was deployed on the real-world UR5 directly without fine-tuning. We use the Aruco markers and depth camera to get the object’s pose. We compared PEBBLE, MRN, RUNE, M-RUNE, QPA and M-SENIOR, selected the optimal policy among all random seeds, and repeated the experiment 20 times. The results are shown in Table \ref{tab:real_experiment}. M-SENIOR achieved the highest success rate on all tasks, indicating that our method is more applicable regarding the noise of real-world input when transferring from simulation to real.

\begin{table}[htbp]
    \centering
    \renewcommand{\arraystretch}{1.05} % 调整行距，使表格更易读
    \setlength{\tabcolsep}{3pt} % 调整列间距
    \caption{Success rate of simulation and real experiments on four tasks}
    \begin{tabular}{lrrrrrrrr}
    \toprule
    \multirow{2}{*}{Method} & 
    \multicolumn{2}{c}{\makecell{{Door Open}}} & 
    \multicolumn{2}{c}{\makecell{{Door Close}}} & 
    \multicolumn{2}{c}{\makecell{{Box Open}}} & 
    \multicolumn{2}{c}{\makecell{{Box Close}}} \\
    \cmidrule(lr){2-3} \cmidrule(lr){4-5} \cmidrule(lr){6-7} \cmidrule(lr){8-9} 
    & sim & real & sim & real & sim & real & sim & real \\
    \midrule
    PEBBLE       & 95\% & 50\% & 95\% & 45\% & 80\% & 30\% & 50\% & 10\% \\
    MRN       & 95\% & 55\% & 95\% & 80\% & 85\% & 35\% & 60\% & 30\%  \\
    RUNE      & 100\% & 40\% & 100\% & 65\% & 80\% & 40\% & 60\% & 30\%  \\
    M-RUNE   & 100\% & 40\% & 100\% & 100\% & 90\% & 55\% & 65\% & 35\% \\
    QPA   & 100\% & 30\% & 100\% & 95\% & 90\% & 35\% & 80\% & 35\% \\
    \addlinespace[0.1em] \midrule
    \textbf{M-SENIOR} & \textbf{100\%} & \textbf{85\%} & \textbf{100\%} & \textbf{100\%} & \textbf{90\%} & \textbf{90\%} & \textbf{100\%}& \textbf{60\%} \\
    \bottomrule
    \end{tabular}
    \label{tab:real_experiment}
\end{table}

\section{CONCLUSIONS}

In this paper, we propose SENIOR, a novel efficient query selection and preference-guided exploration method to improve the feedback- and exploration-efficiency in PbRL. The method combines a motion-distinction-based selection scheme with a preference-guided intrinsic exploration mechanism to efficiently select meaningful and easily comparable segment pairs for high-quality feedback labels and encourage agents to efficiently explore the state space of human interest. The synergy between the two parts accelerates reward learning and policy convergence. We demonstrate the promising experimental results in six simulated robot manipulation tasks and four real-world tasks. In future work, we will investigate how to utilize prior knowledge of large language models (LLM) to perform RL tasks better and further reduce human effort in PbRL. 

% \addtolength{\textheight}{-5cm}   % This command serves to balance the column lengths
%                                   % on the last page of the document manually. It shortens
%                                   % the textheight of the last page by a suitable amount.
%                                   % This command does not take effect until the next page
%                                   % so it should come on the page before the last. Make
%                                   % sure that you do not shorten the textheight too much.

\section*{APPENDIX}

\subsection{SENIOR Algorithm}
This section provides the full procedure for SENIOR in Algorithm \ref{SENIOR}. 
\begin{algorithm}[htbp]
\caption{SENIOR (MRN based)}
\label{SENIOR}
\begin{algorithmic}[1]
    \REQUIRE preference feedback frequency $K$, bi-level optimization frequency $N$, curiosity buffer updating frequency $Q$, number of preference queries per session $M$, curiosity buffer capacity $B$ \COMMENT{both $K$ and $N$ are divisible by $Q$}
    \STATE Initialize $\phi$, $\psi$, and $\theta$ \COMMENT{policy, reward function and Q function network parameter}
    \STATE Initialize preference dataset $\mathcal{D}\gets\emptyset$, replay buffer $\mathcal{B}\gets\emptyset$ and curiosity buffer $\mathcal{B}_{cur}$
    \STATE Update $\phi$ and $\mathcal{B}$ by unsupervised exploration
    \FOR{each iteration}
    \STATE Obtain $\mathbf{s}_{t+1}$ by taking $\mathbf{a}_t\sim\pi_{\phi}(\mathbf{a}_t|\mathbf{s}_t)$
    \STATE Store transitions $\mathcal{B}\gets\mathcal{B}\cup\left\{(\mathbf{s}_t, \mathbf{a}_t, \mathbf{s}_{t+1}, \hat{r}_{\psi}(\mathbf{s}_t, \mathbf{a}_t))\right\}$
    \IF{iteration $\%$ $K == 0$}
        \STATE Generate $\left\{(\sigma^0, \sigma^1)_j\right\}_{j=1}^{M}\sim\mathtt{SAMPLE(\,)}$ (See IV.A)
        \STATE Query preference $(y_j)_{j=1}^{M}$
        \STATE Store preference $\mathcal{D}\gets\mathcal{D}\cup\left\{(\sigma^0, \sigma^1, y)_j\right\}_{j=1}^{M}$
        \STATE Optimize $(\ref{eq:cross-entropy})$ with respect to $\psi$
        \STATE Relabel replay buffer $\mathcal{B}$ using updated $\hat{r}_{\psi}$
        \STATE Update preference KDE $\hat{f}_{\mathcal{P}}$ according to \eqref{eq:f}
    \ENDIF    
    \IF{iteration $\%$ $N == 0$}
        \STATE Sample minibatch from replay buffer $\mathcal{B}$
        \STATE Update $\psi$ with bi-level optimization
        \STATE Relabel replay buffer $\mathcal{B}$ using updated $\hat{r}_{\psi}$
    \ENDIF
    \IF{iteration $\%$ $Q == 0$}
        \STATE \small{Sample minibatch from replay buffer into curiosity buffer} $\mathcal{B}_{cur}\gets\left\{(\mathbf{s}_j, \mathbf{a}_j, \mathbf{s}_{j+1}, \hat{r}_{\psi}(\mathbf{s}_j, \mathbf{a}_j))\right\}_{j=1}^B\sim\mathcal{B}$
        \STATE Update exploration KDE $\hat{f}_{\mathcal{E}}$ and corresponding intrinsic reward $r_{int}$ according to $(\ref{eq:r_int})$
        \STATE Update Curiosity reward of $\mathcal{B}_{cur}$ according to $(\ref{eq:reward_cur})$
    \ENDIF
    \STATE Sample minibatch in proportion from $\mathcal{B}$ and $\mathcal{B}_{cur}$
    \STATE Update $\theta$ and $\phi$ with bi-level optimization.
    \ENDFOR
\end{algorithmic}
\end{algorithm}

\subsection{Experimental Details}
The parameters of SENIOR, PEBBLE, QPA, SAC are shown in Table \ref{hyper of SENIOR}, \ref{hyper of pebble}, \ref{hyper of qpa}, \ref{hyper of sac}. The basic parameters of RUNE and MRN are the same as PEBBLE and are all with Disagreement-based query selection. Except that, in RUNE, $\beta_0=0.05$ and $\rho\in\left\{0.001, 0.0001, 0.00001\right\}$, in MRN the bi-level optimization of frequency is set to 10000.
\begin{table}[!htbp]
    \centering
    \caption{Hyperparameters of SENIOR}
    \begin{tabular}{ll}
    \toprule
    \textbf{Hyperparameter} & \textbf{Value} \\
    \midrule
    Motion selection ratio & 30\% \\
    \makecell[l]{Curiosity buffer updating frequency} & \makecell[l]{1000 (Door Lock,\\Door Open,\\Handle Press)\\500 (Door Unlock,\\Window Close,\\Window Open)} \\
    \makecell[l]{Curiosity buffer and \\Preference
    buffer capacity} & 1000 \\
    Hybrid experience ratio & 30\% \\
    \bottomrule
    \end{tabular}
    \label{hyper of SENIOR}
\end{table}
\begin{table}[!htbp]
    \centering
    \caption{Hyperparameters of PEBBLE}
    \begin{tabular}{ll}
    \toprule
    \textbf{Hyperparameter} & \textbf{Value} \\
    \midrule
    Length of segment & 50 \\
    Frequency of feedback & 5000 \\
    Query selection scheme & Entropy-based \\
    Number of reward functions & 3 \\
    \makecell[l]{Reward model layers /\\hidden units} & 3 / 256 \\
    \makecell[l]{Reward model activation /\\output activation function} & LeakyRelu / TanH \\
   \makecell[l]{ Amount of feedback /\\ feedback amount per session} & \makecell[l]{1000/20 (Door Unlock)\\1000/10 (Door Open)\\250/10 (Handle Press,\\Window Open)\\250/5 (Door Lock,\\Window Close)} \\
    \bottomrule
    \end{tabular}
    \label{hyper of pebble}
\end{table}
\begin{table}[!htbp]
    \centering
    \caption{Hyperparameters of QPA}
    \begin{tabular}{ll}
    \toprule
    \textbf{Hyperparameter} & \textbf{Value} \\
    \midrule
    \makecell[l]{Size of policy-aligned buffer} & \makecell[l]{60(Door Open)\\30(Other tasks)}\\
    Data augmentation ratio $\tau$ & 20 \\
    Hybrid experience replay sample ratio $\omega$ & 0.5 \\
    Min/Max length of subsampled snippets & [35, 45] \\
    \bottomrule
    \end{tabular}
    \label{hyper of qpa}
\end{table}
\begin{table}[!htbp]
    \centering
    \caption{Hyperparameters of SAC}
    \begin{tabular}{ll|ll}
    \toprule
    \textbf{Hyperparameter} & \textbf{Value} & \textbf{Hyperparameter} & \textbf{Value} \\
    \midrule
    Number of layers & 3 & \makecell[l]{Critic target update freq} & 2 \\
    \makecell[l]{Hidden units per\\each layer} & 256 & Optimizer & Adam \\
    Learning rate & 0.0003 & $(\beta_1, \beta_2)$ & (.9, .999) \\
    Batch size & 512 & Critic EMA $\tau$ & 0.005 \\
    Initial temperature & 0.1 & Discount $\gamma$ & .99 \\
    Steps of pre-training & 9000 \\
    \bottomrule
    \end{tabular}
    \label{hyper of sac}
\end{table}
%%%%%%%%%%%%%%%%%%%%%%%%%%%%%%%%%%%%%%%%%%%%%%%%%%%%%%%%%%%%%%%%%%%%%%%%%%%%%%%%
\bibliographystyle{IEEEtran}
\bibliography{IEEEabrv,reference}

\end{document}